\documentclass[10pt,twocolumn,letterpaper]{article}

\usepackage{iccv}
\usepackage{times}
\usepackage{epsfig}
\usepackage{graphicx}
\usepackage{amsmath}
\usepackage{amssymb}
\usepackage{floatrow}
\usepackage{multirow}
\usepackage{threeparttable}
\usepackage{algorithm}
\usepackage{algorithmicx}
\usepackage{algpseudocode}
\usepackage{amsmath}
\usepackage{bm}
\usepackage{times}
\usepackage{epsfig}
\usepackage{graphicx}
\usepackage{amsmath}
\usepackage{amssymb}
\usepackage{algorithm}
\usepackage{algorithmicx}
\usepackage{algpseudocode}
\usepackage{amsmath}
\usepackage{setspace}
\usepackage{color}
\usepackage{amssymb}

\usepackage[breaklinks=true,bookmarks=false]{hyperref}

\iccvfinalcopy 

\usepackage[symbol]{footmisc}

\begin{document}

\title{UVA: A Universal Variational Framework for Continuous Age Analysis}

\author{Peipei Li\textsuperscript{*}, Huaibo Huang\textsuperscript{*}, Yibo Hu, Xiang Wu, Ran He and Zhenan Sun \\
National Laboratory of Pattern Recognition, CASIA, Beijing, China 100190 \\
Center for Research on Intelligent Perception and Computing, CASIA, Beijing, China 100190 \\
University of Chinese Academy of Sciences, Beijing, China 100049\\
Email: \{peipei.li, huaibo.huang, yibo.hu\}@cripac.ia.ac.cn, alfredxiangwu@gmail.com\\ \{rhe, znsun\}@nlpr.ia.ac.cn
}

\maketitle

\footnotetext[1]{These authors have contributed equally to this work.}

\begin{abstract}

Conventional methods for facial age analysis tend to utilize accurate age labels in a supervised way. However, existing age datasets lies in a limited range of ages, leading to a long-tailed distribution. To alleviate the problem, this paper proposes a Universal Variational Aging (UVA) framework to formulate facial age priors in a disentangling manner. Benefiting from the variational evidence lower bound, the facial images are encoded and disentangled into an age-irrelevant distribution and an age-related distribution in the latent space. A conditional introspective adversarial learning mechanism is introduced to boost the image quality. In this way, when manipulating the age-related distribution, UVA can achieve age translation with arbitrary ages. Further, by sampling noise from the age-irrelevant distribution, we can generate photorealistic facial images with a specific age. Moreover, given an input face image, the mean value of age-related distribution can be treated as an age estimator. These indicate that UVA can efficiently and accurately estimate the age-related distribution by a disentangling manner, even if the training dataset performs a long-tailed age distribution. UVA is the first attempt to achieve facial age analysis tasks, including age translation, age generation and age estimation, in a universal framework. The qualitative and quantitative experiments demonstrate the superiority of UVA on five popular datasets, including CACD2000, Morph, UTKFace, MegaAge-Asian and FG-NET.

\end{abstract}

\section{Introduction}

\begin{figure}[t]
\setlength{\abovecaptionskip}{0cm}
\begin{center}
\includegraphics[width=0.75\linewidth]{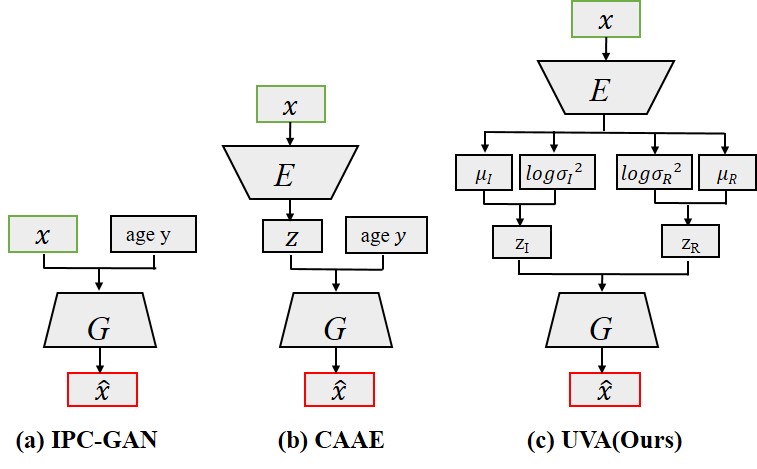}
\end{center}
\vspace{-0.3cm}
   \caption{Compared with the previous conditional generative models \cite{wang2018face,zhang2017age}, UVA learns disentangled age-related and age-irrelevant distributions, which is more suitable for the long-tailed data.}\label{fig:compare}
   \vspace{-0.3cm}
\end{figure}

Facial age analysis, including age translation, age generation and age estimation, is one of crucial components in modern face analysis for entertainment and forensics. Age translation (also known as face aging) aims to aesthetically render the facial appearance to a given age. In recent years, with the developments of generative adversarial network (GAN)~\cite{goodfellow2014generative} and image-to-image translation \cite{Isola2017ImagetoImageTW, Zhu2017UnpairedIT}, impressive progresses \cite{zhang2017age,yang2018learning,wang2018face,li2018global,li2018global1} have been achieved on age translation. These methods often utilize a target age vector as a condition to dominate the facial appearance translation. Fig.~\ref{fig:compare} summarizes the commonly used frameworks for age translation. As shown in Fig.~\ref{fig:compare} (a), IPC-GAN~\cite{wang2018face} directly incorporates the target age with inputs to synthesize facial aging images, while CAAE~\cite{zhang2017age} (shown in Fig.~\ref{fig:compare} (b)) guides age translation by concatenating the age label with the facial image representations in the latent space. It is obvious that the performance of face aging depends on the accurate age labels. Although previous
methods~\cite{yang2018learning,li2018global,li2018global1} have achieved remarkable visual results, in practice, it is difficult to collect labeled images of continuous ages for the intensive aging progression. Since all the existing datasets perform a long-tailed age distribution, researchers often employ the time span of 10 years as the age clusters for age translation. This age cluster formulation potentially limits the diversity of aging patterns, especially for the younger.
\begin{figure}[t]
\setlength{\abovecaptionskip}{0cm}
\begin{center}
\includegraphics[width=1\linewidth]{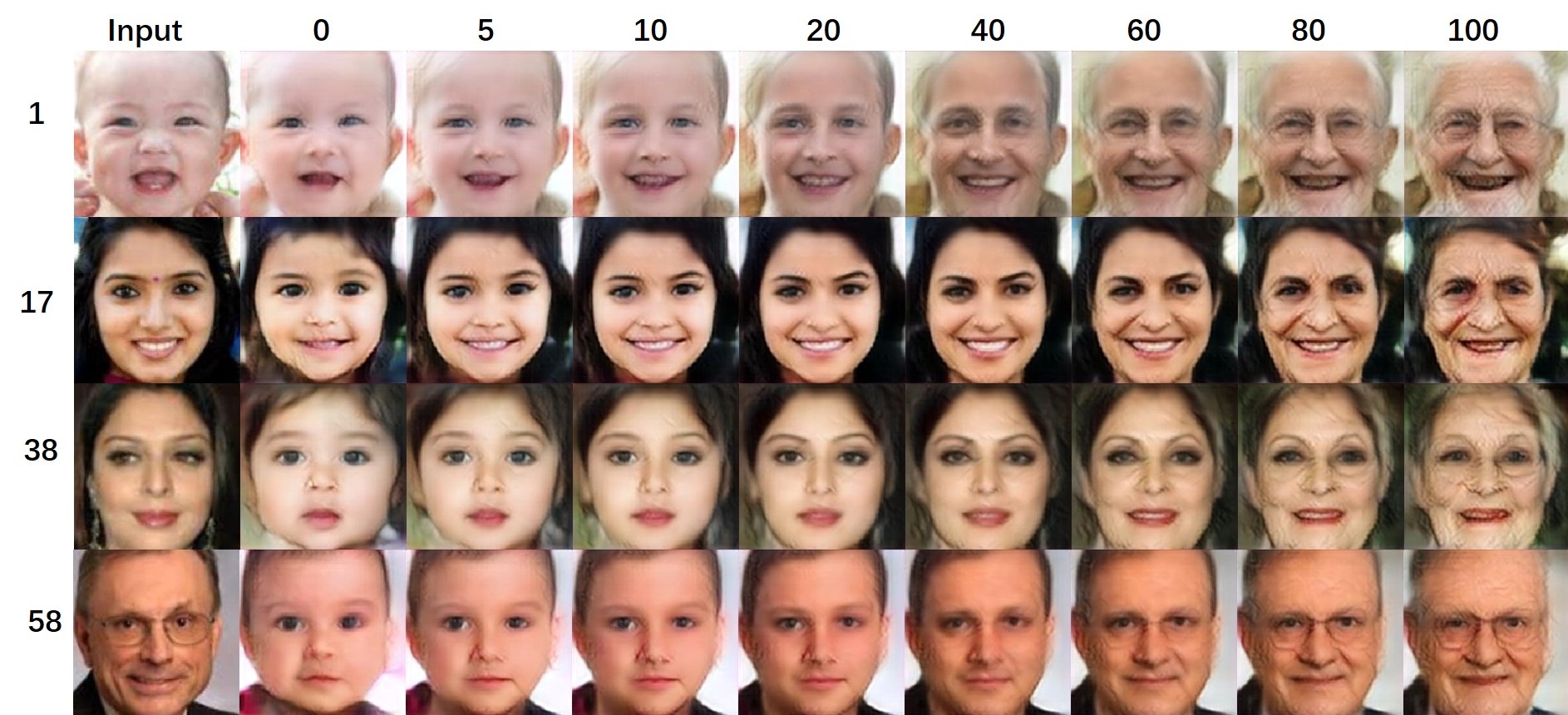}
\end{center}
\vspace{-0.3cm}
   \caption{Continuous age translation results on UTKFace. The first column are the inputs and the rest columns are the synthesized faces from 0 to 100 years old. Note that there are only 8 images at age 100 in UTKFace. }\label{fig:example_translation}
   \vspace{-0.3cm}
\end{figure}

Recently, the variational autoencoder (VAE)~\cite{kingma2013auto} shows the promising ability in discovering interpretability of the latent factors~\cite{burgess2018understanding,huang2018introvae,XWu:2019}. By augmenting VAE with a hyper-parameter $\beta$, $\beta$-VAE~\cite{Matthey2016vaeLB} controls the degree of disentanglement in the latent space. Benefiting from the disentangling abilities in VAE, this paper proposes a novel Universal Variational Aging (UVA) framework to formulate facial age priors in a disentangling manner. Compared with the existing methods, UVA is more capable of disentangling the facial images into an age-related distribution and an age-irrelevant distribution in the latent space, rather than directly utilizing age labels as conditions for age translation. To be specific, the proposed method introduces two latent variables to model the age-related and age-irrelevant information, and employs the variational evidence lower bound (ELBO) to encourage these two parts to be disentangled in the latent space. As shown in Fig. \ref{fig:compare} (c), the age-related distribution is assumed as a Gaussian distribution, where the mean value $\mu_R$ is the real age of the input image, while the age-irrelevant prior is set to a normal distribution $\mathbb{N}(\bf{0},\bf{I})$. The disentangling manner makes UVA perform more flexible and controllable for facial age analysis. Additionally, to synthesize photorealistic facial images, an extended conditional version of introspective adversarial learning mechanism~\cite{huang2018introvae} is introduced to UVA, which self-estimates the differences between the real and generated samples.

   \vspace{-0.1cm}
In this way, by manipulating the mean value $\mu_R$ of age-related distribution, UVA can easily realize facial age translation with arbitrary age (shown in Fig.~\ref{fig:example_translation}), whether the age label exists or not in the training dataset. We furter observe an interesting phenomenon that when sampling noise from the age-irrelevant distribution, UVA can generate photorealistic face images with a specific age. Moreover, given a face image as input, we can easily obtain its age label from the mean value of the age-related distribution, which indicates the ability of UVA to achieve age estimation. As stated above, we can implement three different tasks for facial age analysis in a unified framework. To the best of our knowledge, UVA is the first attempt to achieve facial age analysis, including age translation, age generation and age estimation, in a universal framework. The main contributions of UVA are as follows:

\begin{itemize}
\item We propose a novel Universal Variational Aging (UVA) framework to formulate the continuous facial aging mechanism in a disentangling manner. It leads to a universal framework for facial age analysis, including age translation, age generation and age estimation.

\item Benefiting from the variational evidence lower bound, in UVA, the facial images are encoded and disentangled into an age-related distribution and an age-irrelevant distribution in the latent space. An extended conditional introspective adversarial learning mechanism is introduced to obtain photorealistic facial image synthesis.

\item Different from the existing conditional age translation methods, which utilize the age labels/clusters as a condition, UVA tries to estimate an age distribution from the long-tailed facial age dataset. This age distribution estimation provides a new condition means to model continuous ages that contributes to the interpretability of the image synthesis for facial age analysis.

\item The qualitative and quantitative experiments demonstrate that UVA successfully formulates the facial age prior in a disentangling manner, obtaining state-of-the-art resuts on the CACD2000 \cite{chen2015face}, Morph \cite{ricanek2006morph}, UTKFace \cite{zhifei2017cvpr}, MegaAge-Asian \cite{zhang2017quantifying} and FG-NET \cite{lanitis2002toward} datasets.
\end{itemize}

\section{Related Work}
\subsection{Variational Autoencoder}

Variational Autoencoder (VAE)~\cite{kingma2013auto,rezende2014stochastic} consists of two networks: an inference network ${q_\phi }\left( {z|x} \right)$ maps the data $x$ to the latent variable $z$, which is assumed as a gaussian distribution, and a generative network ${p_\theta }\left( {x|z} \right)$ reversely maps the latent variable $z$ to the visible data $x$. The object of VAE is to maximize the variational lower bound (or evidence lower bound, ELBO) of $\log {p_\theta }\left( x \right)$:
\begin{equation}
   \vspace{-0.1cm}
\label{eq:vae}
\begin{array}{c}
\log {p_\theta }\left( x \right) \ge {E_{{q_\phi }\left( {z|x} \right)}}\log {p_\theta }\left( {x|z} \right) - {D_{KL}}\left( {{q_\phi }\left( {z|x} \right)||p\left( z \right)} \right)
\end{array}
\end{equation}

\begin{figure*}[t]
\setlength{\abovecaptionskip}{0cm}
\begin{center}
\includegraphics[width=0.8\linewidth]{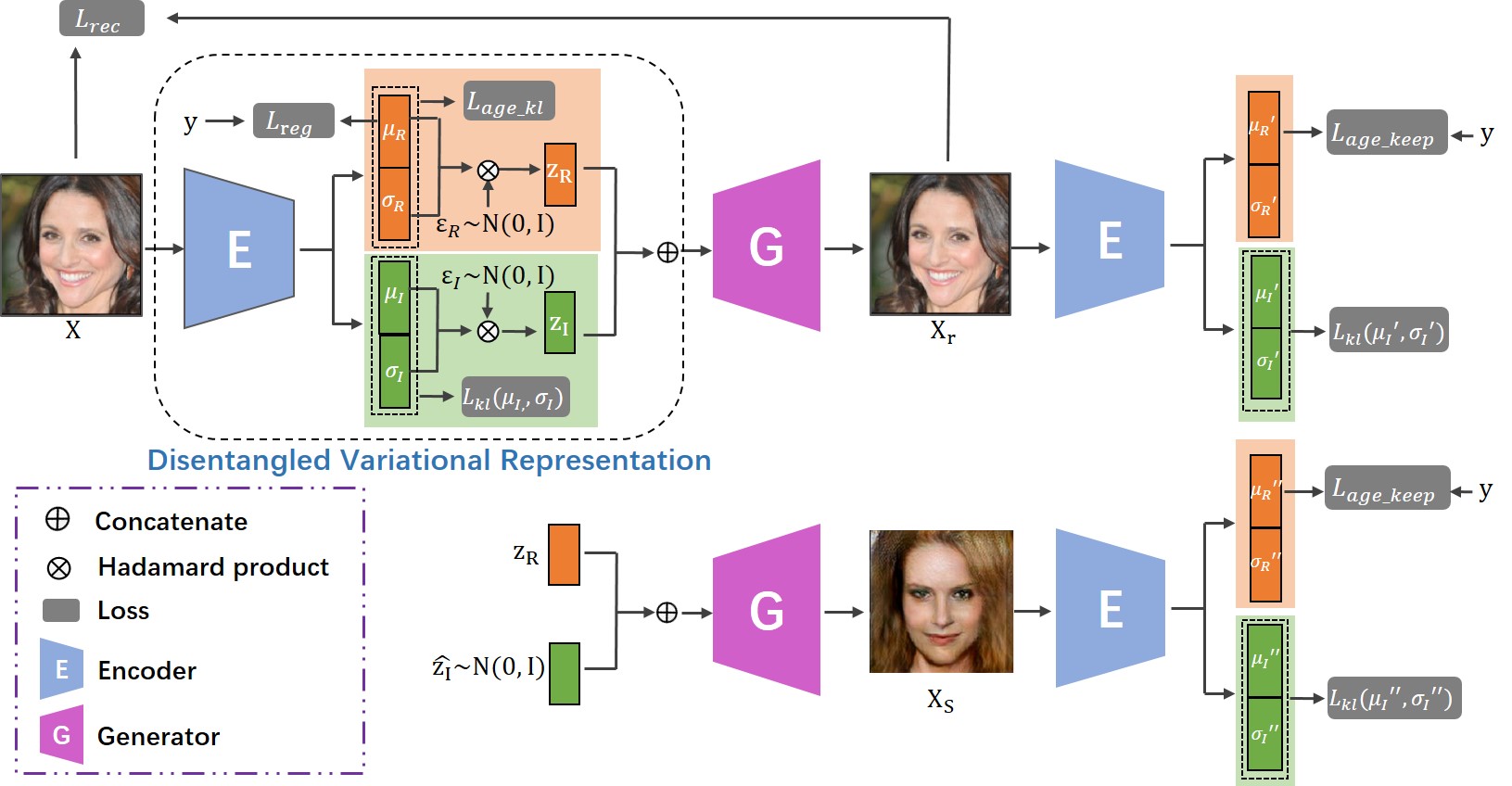}
\end{center}
\vspace{-0.2cm}
   \caption{Overview of the architecture and training flow of our approach. Our model contains two components, the inference network $E$ and the generative network $G$. $X$, $X_r$ and $X_s$ denote the real sample, the reconstruction sample and the new sample, respectively. Please refer to Section 3 for more details.}\label{fig:architecture}
   \vspace{-0.2cm}
\end{figure*}
VAE has shown promising ability to generate complicated data, including faces \cite{huang2018introvae}, natural images \cite{gulrajani2016pixelvae}, text \cite{semeniuta2017hybrid} and segmentation \cite{hou2017deep, sohn2015learning}. Following IntroVAE \cite{huang2018introvae}, we adopt introspective adversarial learning in our method to produce high-quality and realistic face images.

\subsection{Age Translation and Generation}

Recently, deep conditional generative models have shown considerable ability in age translation \cite{zhang2017age, wang2018face, li2018global, li2018global1}. Zhang et al. \cite{zhang2017age} propose a Conditional Adversarial Autoencoder(CAAE) to transform an input facial image to the target age. To capture the rich textures in the local facial parts, Li et al. \cite{li2018global} propose a Global and Local Consistent Age Generative Adversarial Network (GLCA-GAN) method. Meanwhile, Identity-Preserved Conditional Generative Adversarial Networks (IPCGAN)~\cite{wang2018face} introduces an identity-preserved term and an age classification term into age translation. Although these methods have achieved promising visual results, they have limitations in modeling continuous aging mechanism.

With the development of deep generative models, i.e., Generative Adversarial Network(GAN)\cite{goodfellow2014generative}, Variational Autoencoder(VAE), face generation has achieved dramatically success in recent years. Karras et al. \cite{karras2017progressive} provide a new progressive training way for GAN, which grows both the generator and discriminator progressively. Huang et al. \cite{huang2018introvae} propose the Introspective Variational Autoencoders(IntroVAE), in which the inference model and the generator are jointly trained in an introspective way. 
\section{Approach}

In this section, we propose a universal variational aging framework for age translation, age generation and age estimation.
The key idea is to employ an inference network $E$ to embed the facial image into two disentangled variational representations, where one is age-related and another is age-irrelevant, and a generator network $G$ to produce photo-realistic images from the re-sampled latent representations.
As depicted in Fig.~\ref{fig:architecture}, we assign two different priors to regularize the inferred representations, and train $E$ and $G$ with age preserving regularization in an introspective adversarial manner. The details are given in the following.

\subsection{Disentangled Variational Representations}

In the original VAE~\cite{kingma2013auto}, a probabilistic latent variable model is learnt by maximizing the variational lower bound to the marginal likelihood of the observable variables. However, the latent variable $z$ is difficult to interpret and control, for each element of $z$ is treated equally in the training. To alleviate this, we manually split $z$ into two parts, i.e., one part $z_R$ representing the age-related information and another part $z_I$ representing the age-irrelevant information.

Assume $z_R$ and $z_I$  are independent on each other, then the posterior distribution ${q_\phi }\left( {z|x} \right) = {q_\phi }\left( {z_R|x} \right) {q_\phi }\left( {z_I|x} \right)$. The prior distribution $p\left( z \right) = {p_R}\left( z_R \right)  {p_I}\left( z_I \right) $, where ${p_R}\left( z_R \right)$  and ${p_I}\left( z_I \right)$ are the prior distributions for $z_R$ and $z_I$, respectively. According to Eq.~(\ref{eq:vae}), the optimization objective for the modified VAE is to maximize the lower bound(or evidence lower bound, ELBO) of $\log {p_\theta }\left( x \right)$:
\begin{equation}
   \vspace{-0.2cm}
\label{eq:vae:ours}
\begin{split}
\log {p_\theta }\left( x \right) \ge &{E_{{q_\phi }\left( {z_R, z_I|x} \right)}}\log {p_\theta }\left( {x|z_R, z_I} \right) \\
 &- {D_{KL}}\left( {{q_\phi }\left( {z_R|x} \right)||p_R\left( z_R \right)} \right) \\
  &- {D_{KL}}\left( {{q_\phi }\left( {z_I|x} \right)||p_I\left( z_I \right)} \right),
\end{split}
   \vspace{-0.3cm}
\end{equation}

To make $z_R$ and $z_I$  correspond with different types of facial information, ${p_R}\left( z_R \right)$  and ${p_I}\left( z_I \right)$ are set to be different distributions. They are both centered isotropic multivariate Gaussian but
 ${p_R}\left( z_R \right) = \mathbb{N}\left( {{\bf{y}},{\bf{I}}} \right)$ and
${p_I}\left( z_I \right) = \mathbb{N}\left( {{\bf{0}},{\bf{I}}} \right)$,
where $\textbf{y}$ is a vector filled by the age label $y$ of $x$.
Through using the two age-related and age-irrelevant priors, maximizing the above variational lower bound encourages
the posterior distribution
${q_\phi }\left( {z_R|x} \right)$ and ${q_\phi }\left( {z_I|x} \right)$
to be disentangled and to model the age-related and age-irrelevant information, respectively.

Following the original VAE~\cite{kingma2013auto}, we assume the posterior ${q_\phi }\left( {z_R|x} \right)$ and ${q_\phi }\left( {z_I|x} \right)$ follows two centered isotropic multivariate Gaussian, respectively, i.e.,
${q_\phi }\left( {z_R|x} \right) = \mathbb{N}\left( z_R; \mu_R, \sigma_R^2 \right)$, and
${q_\phi }\left( {z_I|x} \right) = \mathbb{N}\left( z_I; \mu_I, \sigma_I^2 \right)$.
As depicted in Fig.~\ref{fig:architecture}, $\mu_R$, $\sigma_R$, $\mu_I$ and $\sigma_I$ are the output vectors of the inference network $E$. The input $z$ of the generator network $G$ is the concatenation of $z_R$ and $z_I$, where $z_R$ and $z_I$ are sampled from $\mathbb{N}\left( z_R; \mu_R, \sigma_R^2 \right)$ and $\mathbb{N}\left( z_I; \mu_I, \sigma_I^2 \right)$ using a reparameterization trick, respectively, i.e., ${z_R} = {\mu _R} + \epsilon_R \odot \sigma _R$, ${z_I} = {\mu _I} + \epsilon_I \odot \sigma _I$, where $\epsilon_R \sim \mathbb{N}\left( {{\bf{0}},{\bf{I}}} \right)$,
$\epsilon_I \sim \mathbb{N}\left( {{\bf{0}},{\bf{I}}} \right)$.

For description convenience, the optimization object in Eq.~(\ref{eq:vae:ours}) is rewritten in the negative version:
\begin{equation}
   \vspace{-0.15cm}
\label{eq:vae:ours:neg}
L_{ELBO} = L_{rec} + L_{age\_kl} + L_{kl} \left(\mu_I, \sigma_I \right)
   \vspace{-0.1cm}
\end{equation}
where $L_{rec}$, $L_{age\_kl}$ and $L_{kl} \left(\mu_I, \sigma_I \right)$
denote the three terms in Eq.~(\ref{eq:vae:ours}), respectively.
They can be computed as below:
\begin{equation}
   \vspace{-0.15cm}
\label{eq:vae:ours:rec}
L_{rec} = \frac{1}{2} \| x - x_r \|^{2}_F
\end{equation}
   \vspace{-0.3cm}
\begin{equation}
\label{eq:vae:ours:agekl}
L_{age\_kl} = \frac{1}{2} \sum_{i=1}^{C}  ((\mu_{R}^{i} - y)^2 + (\sigma_{R}^{i})^2 - \log ((\sigma_{R}^{i})^2) - 1)
   \vspace{-0.2cm}
\end{equation}
\begin{equation}
\label{eq:vae:ours:kl}
L_{kl} \left(\mu_I, \sigma_I \right) = \frac{1}{2}  \sum_{i=1}^{C}  ( (\mu_{I}^{i})^2 + (\sigma_{I}^{i})^2 - \log ((\sigma_{I}^{i})^2) - 1)
   \vspace{-0.2cm}
\end{equation}
where $x_r$ is the reconstruction image of the input $x$, $y$ is the age label of $x$, $C$ denotes the dimension of $z_R$ and $z_I$.

\subsection{Introspective Adversarial Learning}

To alleviate the problem of generating blurry samples in VAEs, the introspective adversarial learning mechanism~\cite{huang2018introvae} is introduced to the proposed UVA.
This makes the model able to self-estimate the differences between the real and generated samples without extra adversarial discriminators. The inference network $E$ is encouraged to distinguish between the real and generated samples while the generator network $G$ tries to fool it as GANs.

Different from IntroVAE~\cite{huang2018introvae}, the proposed method employs a part of the posterior distribution, rather than the whole distribution, to serve as the estimator of the image reality. Specifically, the age-irrelevant posterior ${q_\phi }\left( {z_I|x} \right)$ is selected to help the adversarial learning. When training $E$, the model minimizes the KL-distance of the posterior ${q_\phi }\left( {z_I|x} \right)$ from its prior $p_I \left( z_I\right)$ for the real data and maximize it for the generated samples. When training $G$, the model minimizes this KL-distance for the generated samples.

Similar to IntroVAE~\cite{huang2018introvae}, the proposed UVA is trained to discriminate the real data from both the model reconstructions and samples. As shown in Fig.~\ref{fig:architecture},  these two types of samples are the reconstruction sample $x_r = G\left(z_R, z_I\right)$ and the new samples $x_s = G\left(z_R, \hat{z}_I\right)$,
where $z_R$, $z_I$ and $\hat{z}_I$ are sampled from $q_{\phi} \left( z_R | x \right)$, $q_{\phi} \left( z_I | x \right)$ and $p_I \left(z_I\right)$, respectively.

The adversarial training objects for $E$ and $G$ are defined as below:
\begin{equation}
   \vspace{-0.2cm}
\label{eq:vae:ours:e}
\begin{split}
L_{adv}^{E} = &L_{kl} \left(\mu_I, \sigma_I \right) + \alpha \{ \left[m - L_{kl} \left(\mu'_I, \sigma'_I \right)  \right]^+ \\
    & + \left[m - L_{kl} \left(\mu''_I, \sigma''_I \right)  \right]^+ \} ,
\end{split}
\end{equation}
\begin{equation}
\label{eq:vae:ours:g}
L_{adv}^{G} = L_{kl} \left(\mu_I', \sigma'_I \right) + L_{kl} \left(\mu_I'', \sigma''_I \right),
   \vspace{-0.1cm}
\end{equation}
where $m$ is a positive margin, $\alpha$ is a weighting coefficient,  $\left(\mu_I, \sigma_I\right)$, $\left(\mu'_I,\sigma'_I\right)$ and $\left(\mu''_I, \sigma''_I\right)$ are computed from the real data $x$, the reconstruction sample $x_r$ and the new samples $x_s$, respectively.

The proposed UVA can be viewed as a conditional version of IntroVAE~\cite{huang2018introvae}. The disentangled variational representation and the modified introspective adversarial learning makes it superior to IntroVAE in the interpretability and controllability of image generation.

\subsection{Age Preserving Regularization}

Age accuracy is important to facial age analysis. The most current face aging methods\cite{zhang2017age,wang2018face,li2018global,li2018global1, yang2018learning} usually utilize an additional pre-trained age estimation network \cite{wu2018light} to supervise the generated face images.
While in our method, facial age can be estimated using the inference network $E$ by computing the mean value of the inferred vector $\mu_R$.
To better capture and disentangle age-related information, an age regularization term is utilized on the learned representation $z_R$:
\begin{equation}
   \vspace{-0.2cm}
\begin{array}{c}
{L_{reg}} = \left| \frac{1}{C}\sum\limits_{c = 1}^{C} {\mu_{R}^{i} - y} \right|
\end{array}
   \vspace{-0.0cm}
\end{equation}
where $C$ denotes the dimension of $z_R$, $\mu_R$ is the output vector of the input $x$, $y$ is the age label of $x$.

For further supervising the generator $G$ to reconstruct the age-related information, a similar age regularization term is also employed on the reconstructed and generated samples in Fig.~\ref{fig:architecture}, i.e., $x_r = G\left(z_R, z_I\right)$ and $x_s = G\left(z_R, \hat{z}_I\right)$. The age preserving loss is reformulated as:
\begin{equation}
   \vspace{-0.1cm}
\begin{array}{c}
{L_{age\_keep}} = \left| \frac{1}{C}\sum\limits_{i = 1}^{C} {{\mu'}_{R}^{i} - y} \right| + \left| \frac{1}{C}\sum\limits_{i = 1}^{C} {{\mu''}_{R}^{i} - y} \right|
\end{array}
   \vspace{-0.1cm}
\end{equation}
where $\mu'$ and $\mu''$ are computed by $E$ from $x_r$ and $x_s$, respectively.

\subsection{Training UVA networks}

As shown in Fig.~\ref{fig:architecture}, the inference network $E$ embeds the facial image $x$ into two disentangled distributions, i.e.,
$\mathbb{N}\left( z_R; \mu_R, \sigma_R^2 \right)$ and $\mathbb{N}\left( z_I; \mu_I, \sigma_I^2 \right)$,
where $(\mu_R, \sigma_R)$ and $(\mu_I, \sigma_I)$ are the output vectors of $E$.
The generator network $G$ produces the reconstruction $x_r = G(z_R, z_I)$ and the sample $x_s = G(z_R, {\hat{z}}_I)$,
where  $z_R$, $z_I$ and $\hat{z}_I$ are sampled from $q_{\phi} \left( z_R | x \right)$, $q_{\phi} \left( z_I | x \right)$ and $p_I \left(z_I\right)$, respectively.
To learn the disentangled variational representations and produce high-fidelity facial images,
the network is trained using a weighted sum of the above losses, defined as:
\begin{equation}
   \vspace{-0.2cm}
\begin{array}{c}
{L^E} = L_{rec} + \lambda_1 L_{age\_kl} + \lambda_2 L_{adv}^{E} + \lambda_3 L_{reg}
\end{array}
\end{equation}
\begin{equation}
\begin{array}{c}
{L^G} = L_{rec} + \lambda_4 L_{adv}^{G} + \lambda_5 L_{age\_keep}
\end{array}
\end{equation}
where $\lambda _1$, $\lambda _2$, $\lambda _3$, $\lambda _4$, $\lambda _5$ are the trade-off parameters to balance the important of losses.
Noted that the third term $L_{kl}(\mu_I, \sigma_I)$ in the ELBO loss in Eq.~(\ref{eq:vae:ours:neg}) has been contained in the adversarial loss for $E$, i.e., the first term of $L_{adv}^{E}$ in Eq.~(\ref{eq:vae:ours:e}).

\subsection{Inference and Sampling}

By regularizing the disentangled representations with the age-related prior $p_R\left( {{z_R}} \right)=\mathbb{N}\left( {{\bf{y}},{\bf{I}}} \right)$ and age-irrelevant prior $p_I\left( {{z_I}} \right)=\mathbb{N}\left( {{\bf{0}},{\bf{I}}} \right)$, our UVA is thus a universal framework for age translation, generation and estimation tasks, as illustrated in Fig. \ref{fig:compare} (c).

\textbf{Age Translation}
To achieve the age translation task, we concatenate the age-irrelevant variable ${z _I}$ and a target age variable ${\hat{z}_R}$ as the input of the generator $G$ ,where ${z _I}$ is sampled from the posterior distribution $q_\phi (z_I|x)$ over the input face $x$ and ${\hat{z}_R}$  is sampled from $p_R\left( {{z_R}} \right)$. The age translation result ${\hat x}$ is written as:
\begin{equation}
   \vspace{-0.3cm}
\begin{array}{c}
\hat{x} = G\left(\hat{z}_R,z_I \right)
\end{array}
   \vspace{-0.15cm}
\end{equation}

\textbf{Age Generation}
For the age generation task, there are two settings. One is to generate image from noise ${\hat{z}_I}$ with any ages ${\hat{z}_R}$. To be specific, we concatenate a target age variable ${\hat{z}_R}$ and an age-irrelevant variable ${\hat{z}_I}$ as the input of $G$, where ${\hat{z}_R}$ and ${\hat{z}_I}$ are sampled from $p_R\left( {{z_R}} \right)$ and $p_I\left( {{z_I}} \right)$, respectively. Then the age generation result is:
\begin{equation}
   \vspace{-0.15cm}
\begin{array}{c}
{{\hat x}} = G\left( \hat{z}_R, \hat{z}_I \right)
\end{array}
   \vspace{-0.15cm}
\end{equation}

The other is to generate image from noise ${\hat{z}_I}$, which shares the same age-related variable ${z_{R}}$ with the input. Specifically, we concatenate the age-related variable ${z _R}$ and an age-irrelevant variable ${\hat{z}_I}$ as the input of the $G$,
where ${z _R}$ is sampled from the posterior distribution $q_\phi (z_R|x)$ over the input face $x$
 and ${\hat{z}_I}$ is sampled from $p\left( {{z_I}} \right)$. The age generation result is formulated as:
\begin{equation}
   \vspace{-0.15cm}
\begin{array}{c}
{\hat x} = G\left(z_R ,\hat{z}_I \right)
\end{array}
   \vspace{-0.15cm}
\end{equation}

\textbf{Age Estimation}  In this paper, age estimation is also conducted by the proposed UVA to verify the age-related variable extracting and disentangling performance. We calculate the mean value of $C$-dimension vector $\mu_R$ as the age estimation result, defined as:
\begin{equation}
   \vspace{-0.15cm}
\begin{array}{c}
\hat{y} = \frac{1}{C} \sum\limits_{i=1}^{C} \mu_R^i
\end{array}
   \vspace{-0.15cm}
\end{equation}
where $\mu_R$ is one of the output vectors of the inference network $E$.

\section{Experiments}
\subsection{Datasets and Settings}
\subsubsection{Datasets}
\textbf{Cross-Age Celebrity Dataset (CACD2000)} \cite{chen2015face} consists of 163,446 color facial images of 2,000 celebrities, where the ages range from 14 to 62 years old. However, there are many dirty data in it, which leads to a challenging model training. We choose the classical 80-20 split on CACD2000.

\textbf{Morph} \cite{ricanek2006morph} is the largest public available dataset collected in the constrained environment. It contains 55,349 color facial images of 13,672 subjects with ages ranging from 16 to 77 years old. Conventional 80-20 split is conducted on Morph.

\textbf{UTKFace} \cite{zhifei2017cvpr} is a large-scale facial age dataset with a long age span, which ranges from 0 to 116 years old. It contains over 20,000 facial images in the wild with large variations in expression, occlusion, resolution and pose. We employ classical 80-20 split on UTKFace.

\textbf{MegaAge-Asian} \cite{zhang2017quantifying} is a newly released facial age dataset, consisting of 40,000 Asian faces with ages ranging from 0 to 70 years old. There are extremely variations, such as distortion, large-area occlusion and heavy makeup in MegaAge-Asian. Following \cite{zhang2017quantifying}, we reserve 3,945 images for testing and the remains are treated as training set.

\textbf{FG-NET} \cite{lanitis2002toward} contains 1,002 facial images of 82 subjects. We employ it as the testing set to evaluate the generalization of UVA.

\subsubsection{Experimental Settings}
Following \cite{li2018global1}, we employ the multi-task cascaded CNN \cite{zhang2016joint} to detect the faces. All the facial images are cropped and aligned into 224 $\times$ 224, while the most existing methods \cite{wang2018face, zhang2017age, li2018global} only achieve age translation with 128 $\times$ 128.  Our model is implemented with Pytorch. During training, we choose Adam optimizer \cite{kingma2014adam} with ${\beta _1}$ of 0.9,  ${\beta _2}$ of 0.99, a fixed learning rate of $2 \times {10^{ - 4}}$ and batch size of 28. The trade-off parameters $\lambda _1$, $\lambda _2$, $\lambda _3$, $\lambda _4$, $\lambda _5$ are all set to 1. Besides, $m$ is set to 1000 and $\alpha$ is set to 0.5. More details of the network architectures and training processes are provided in the supplementary materials.

\subsection{Qualitative Evaluation of UVA}

\begin{figure}[t]
\setlength{\abovecaptionskip}{0cm}
\setlength{\belowcaptionskip}{-0.2cm}
\begin{center}
\includegraphics[width=0.9\linewidth]{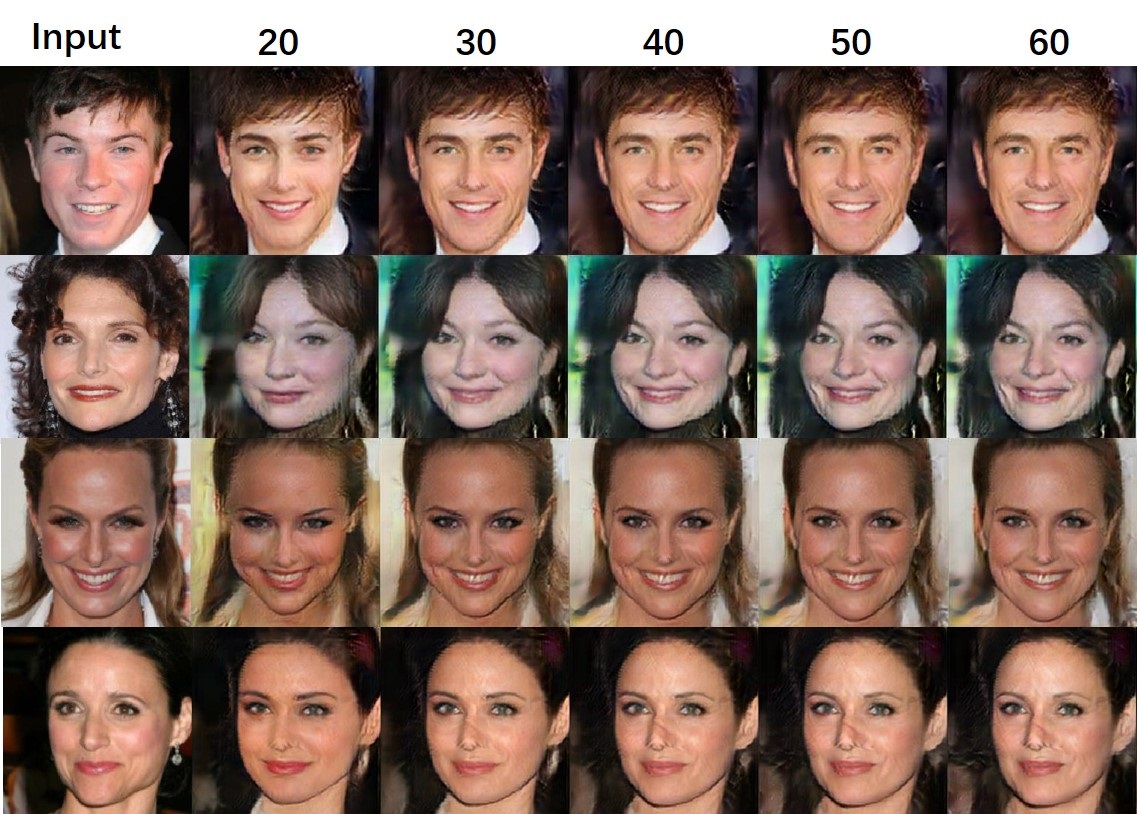}
\end{center}
\vspace{-0.3cm}
   \caption{Age translation results on CACD2000. The first column is the input and the rest five columns are the synthesized results.}\label{fig:cacd2000}
   \vspace{-0.3cm}
\end{figure}

\begin{figure}[t]
\setlength{\abovecaptionskip}{0cm}
\begin{center}
\includegraphics[width=0.9\linewidth]{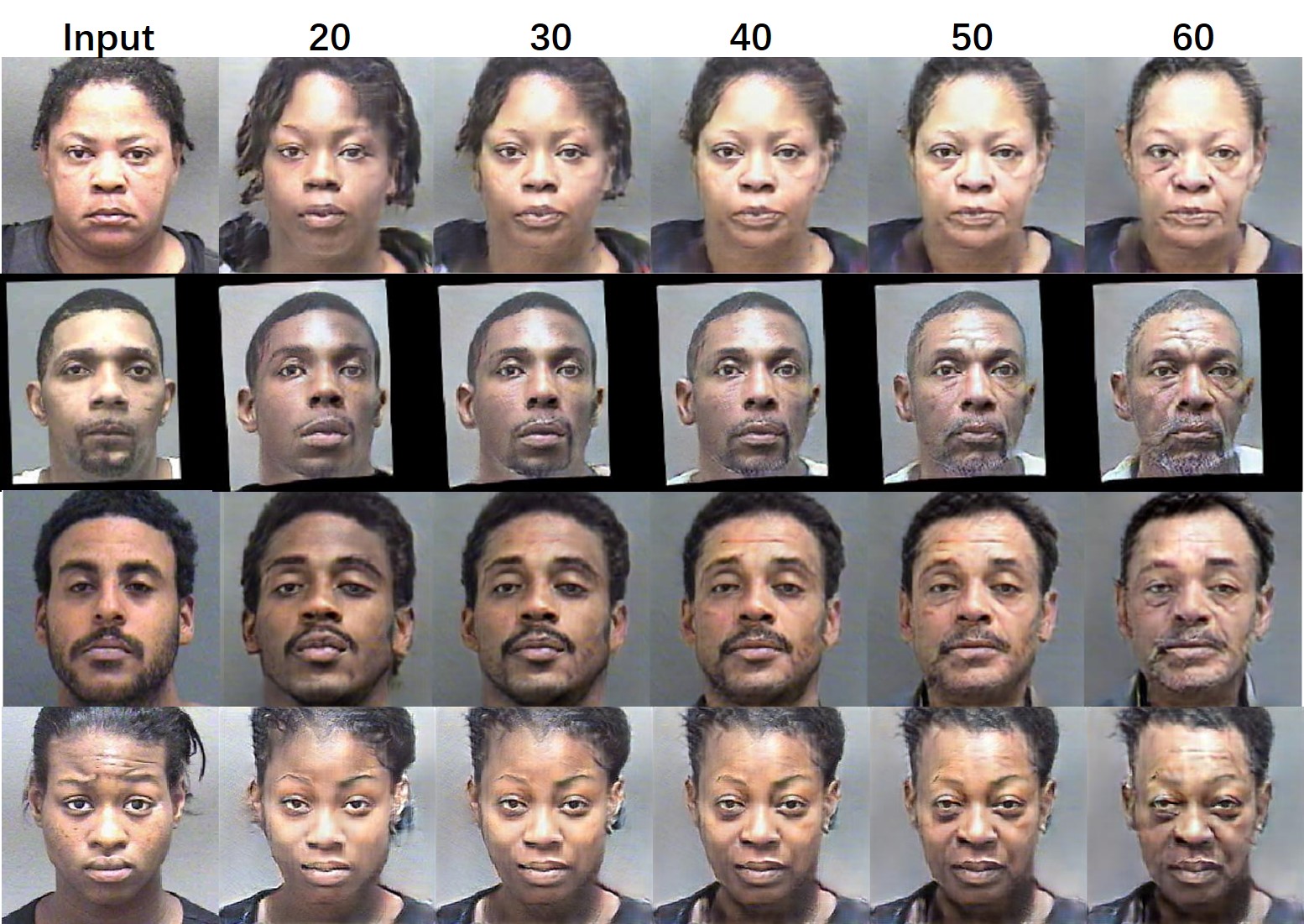}
\end{center}
\vspace{-0.3cm}
   \caption{Age translation results on Morph. The first column is the input and the rest five columns are the synthesized results.}\label{fig:morph}
   \vspace{-0.3cm}
\end{figure}

By manipulating the mean value $\mu_R$ and sampling from age-related distribution, the proposed UVA can translate arbitrary age based on the input. Fig. \ref{fig:cacd2000} and \ref{fig:morph} present the age translation results on CACD2000 and Morph, respectively. We observe that the synthesized faces are getting older and older with ages growing. Specifically, the face contours become longer, the forehead hairline are higher and the nasolabial folds are deepened.
\subsubsection{Age Translation}
\begin{figure}[t]
\setlength{\abovecaptionskip}{0cm}
\setlength{\belowcaptionskip}{-0.2cm}
\begin{center}
\includegraphics[width=0.9\linewidth]{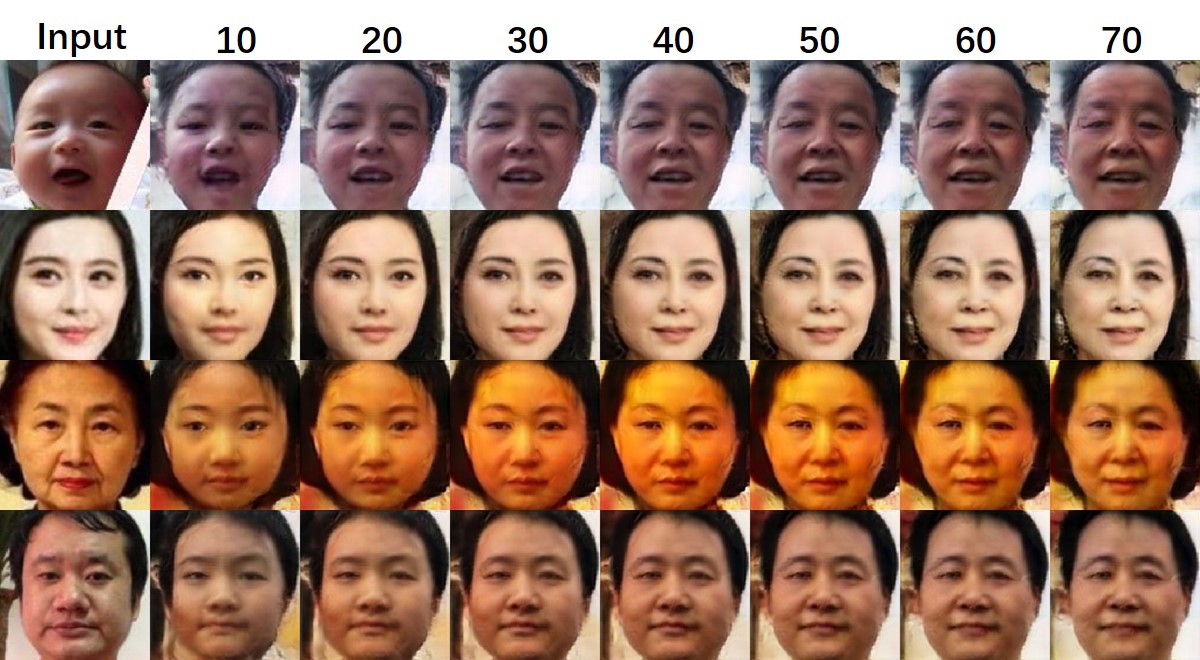}
\end{center}
\vspace{-0.3cm}
   \caption{Age translation results on MegaAge-Asian. The first column is the input and the rest are the synthesized results.}\label{fig:ma}
   \setlength{\belowcaptionskip}{-0.2cm}
   \vspace{-0.2cm}
\end{figure}

\begin{figure}[t]
\vspace{-0.2cm}
\setlength{\abovecaptionskip}{0cm}
\setlength{\belowcaptionskip}{-0.2cm}
\begin{center}
\includegraphics[width=0.85\linewidth]{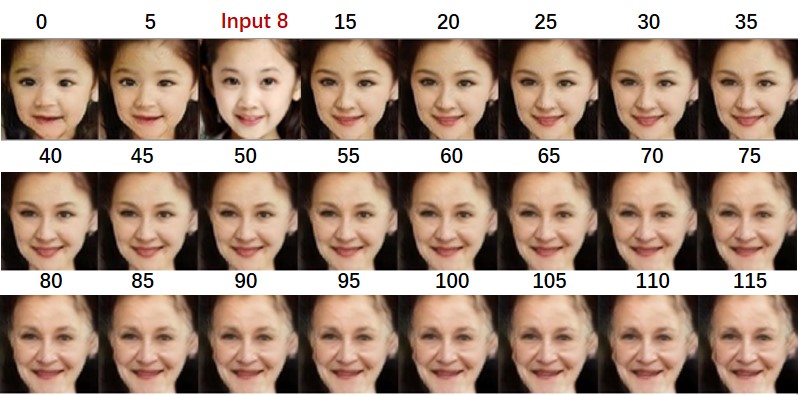}
\end{center}
\vspace{-0.3cm}
   \caption{Age translation results on UTKFace from 0 to 115 years old.}\label{fig:utkface}
    \vspace{-0.3cm}
\end{figure}

Since both CACD2000 and Morph lack of images of children, we conduct age translation on MegaAge-Asian and UTKFace. Fig. \ref{fig:ma} shows the translation results on MegaAge-Asian from 10 to 70 years old. Fig. \ref{fig:utkface} describes the results on UTKFace from 0 to 115 years old with an interval of 5 years old. Obviously, from birth to adulthood, the aging effect is mainly shown on craniofacial growth, while the aging effect from adulthood to elder is reflected on the skin aging, which is consistent with human physiology.

\begin{figure}[t]
\setlength{\abovecaptionskip}{0cm}
\begin{center}
\includegraphics[width=0.8\linewidth]{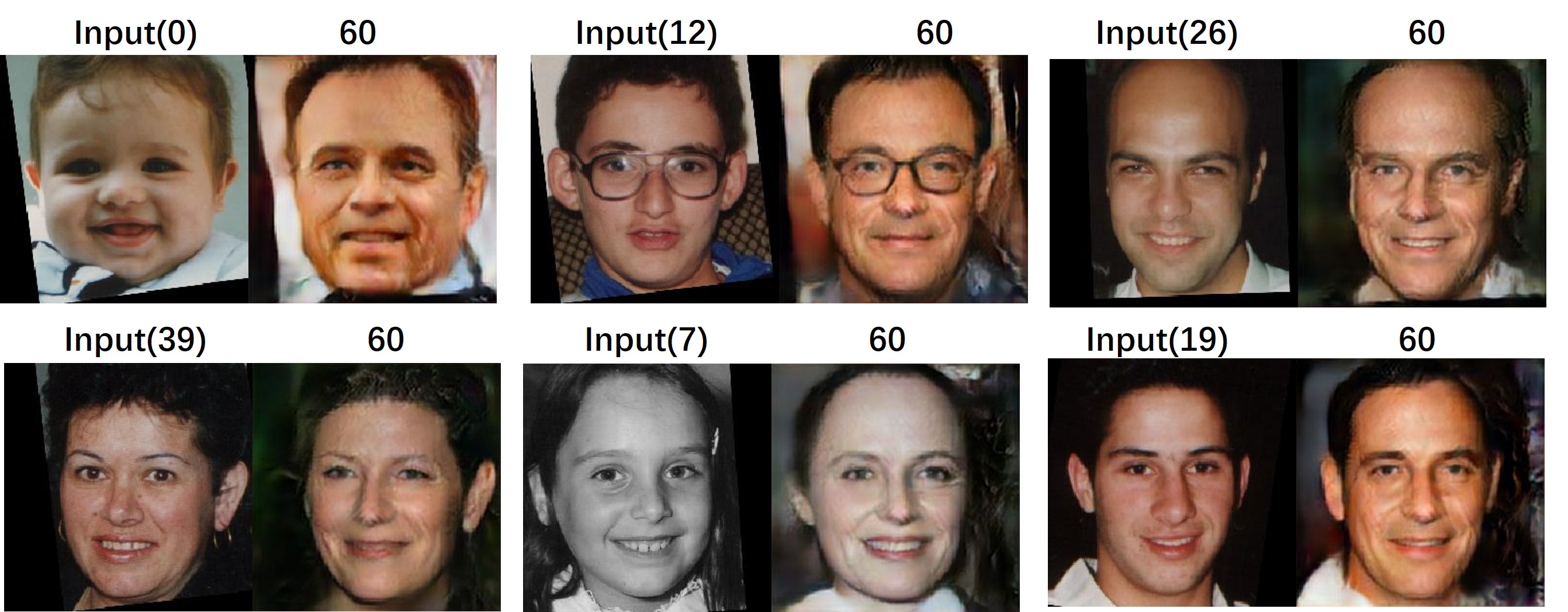}
\end{center}
\vspace{-0.3cm}
   \caption{Cross-dataset age translation results on FG-NET.}\label{fig:fg-net}
 \vspace{-0.3cm}
\end{figure}
\begin{figure}[t]
\vspace{-0.2cm}
\setlength{\abovecaptionskip}{0cm}
\begin{center}
\includegraphics[width=0.85\linewidth]{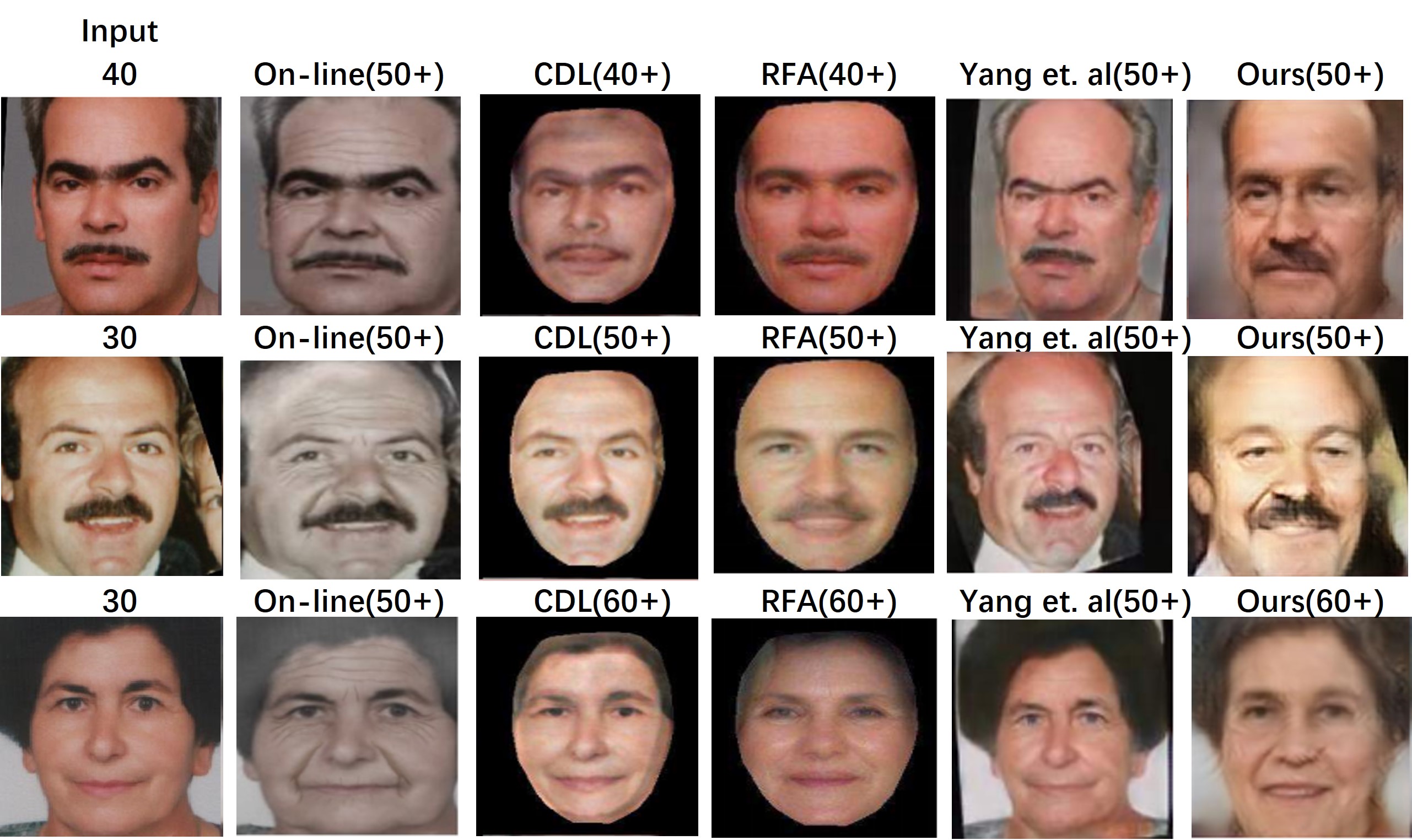}
\end{center}
\vspace{-0.3cm}
   \caption{Comparison with the previous works on FG-NET.}\label{fig:comparison_fgnet}
   \vspace{-0.3cm}
\end{figure}
Furthermore, superior to the most existing methods \cite{zhang2017age, wang2018face, li2018global, li2018global1} that can only generate images at specific age groups, our UVA is able to realize continuous age translation with 1-year-old interval. The translation results on UTKFace from 0 to 119 years old with 1-year-old interval is presented in the supplementary materials, due to the page limitation.

To evaluate the model generalization, we test UVA on FG-NET and show the translation results in Fig. \ref{fig:fg-net}. The left image of each subject is the input and the right one is the translated image with 60 years old. Note that our UVA is trained only on CACD2000 and tested on FG-NET.

Different from the previous age translation works that divide the data into four or nine age groups, our proposed UVA models continuous aging mechanism. The comparison with prior works, including AgingBooth App\cite{AgingBooth}, CDL\cite{shu2015personalized}, RFA\cite{wang2016recurrent} and Yang et al.\cite{yang2017learning}, is depicted in Fig. \ref{fig:comparison_fgnet}.

\subsubsection{Age Generation}
\begin{figure}[t]
\setlength{\abovecaptionskip}{0cm}
\begin{center}
\includegraphics[width=0.9\linewidth]{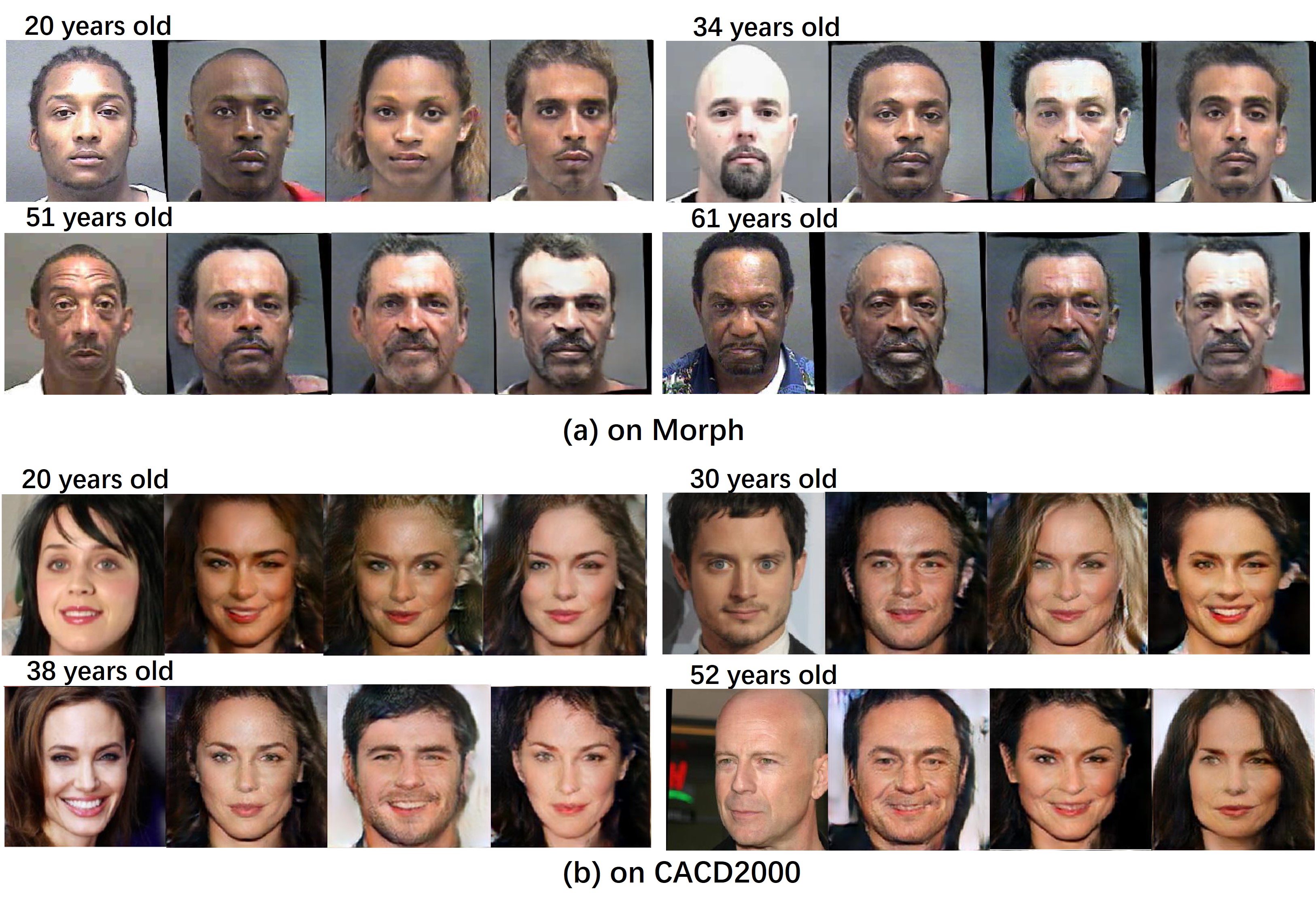}
\end{center}
\vspace{-0.3cm}
   \caption{Age generation results on Morph and CACD2000. For each image group, the first column is the input and the rest three columns are the generated faces with the same age-related representation as the input face.}\label{fig:generation_from_image}
   \vspace{-0.3cm}
\end{figure}

Benefiting from the disentangled age-related and age-irrelevant distributions in the latent space, the proposed UVA is able to realize age generation. On the one hand, conditioned on a given facial image, UVA can generate new faces with the same age-related representation as the given image by fixing $z_R$ and sampling $\hat{z}_I$. Fig. \ref{fig:generation_from_image} presents some results under this situation on the Morph and CACD2000 datasets. Our UVA generates diverse faces (such as different genders, appearances and haircuts) with the same age as the input facial image.

\begin{figure}[t]
\setlength{\abovecaptionskip}{0cm}
\setlength{\belowcaptionskip}{-0.2cm}
\begin{center}
\includegraphics[width=0.75\linewidth]{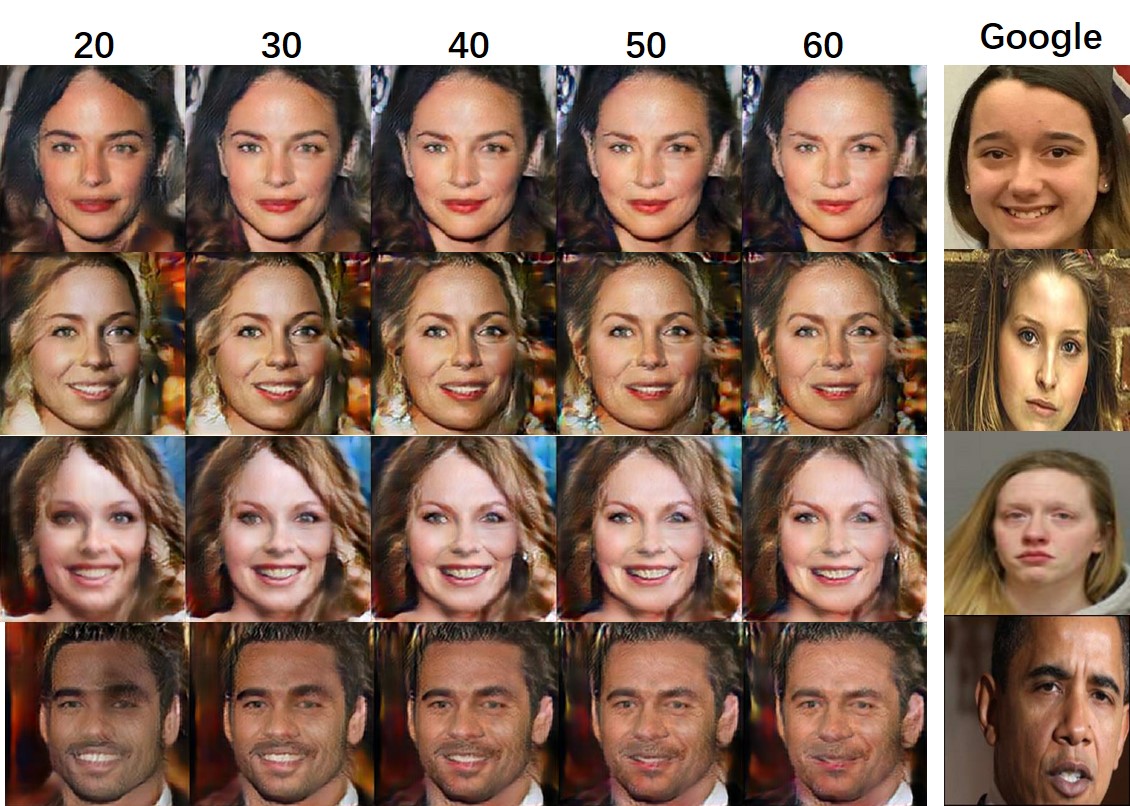}
\end{center}
\vspace{-0.3cm}
   \caption{Age generation results on CACD2000. In the left, each row has the same age-irrelevant representation and each column has the same age-related representation. The right shows the most similar image from google for the generated image in the left. (All of the four subjects are generated from noise.)}\label{fig:cacd2000_generation}
   \vspace{-0.3cm}
\end{figure}

\begin{figure}[t]
\setlength{\abovecaptionskip}{0cm}
\setlength{\belowcaptionskip}{-0.2cm}
\begin{center}
\includegraphics[width=0.75\linewidth]{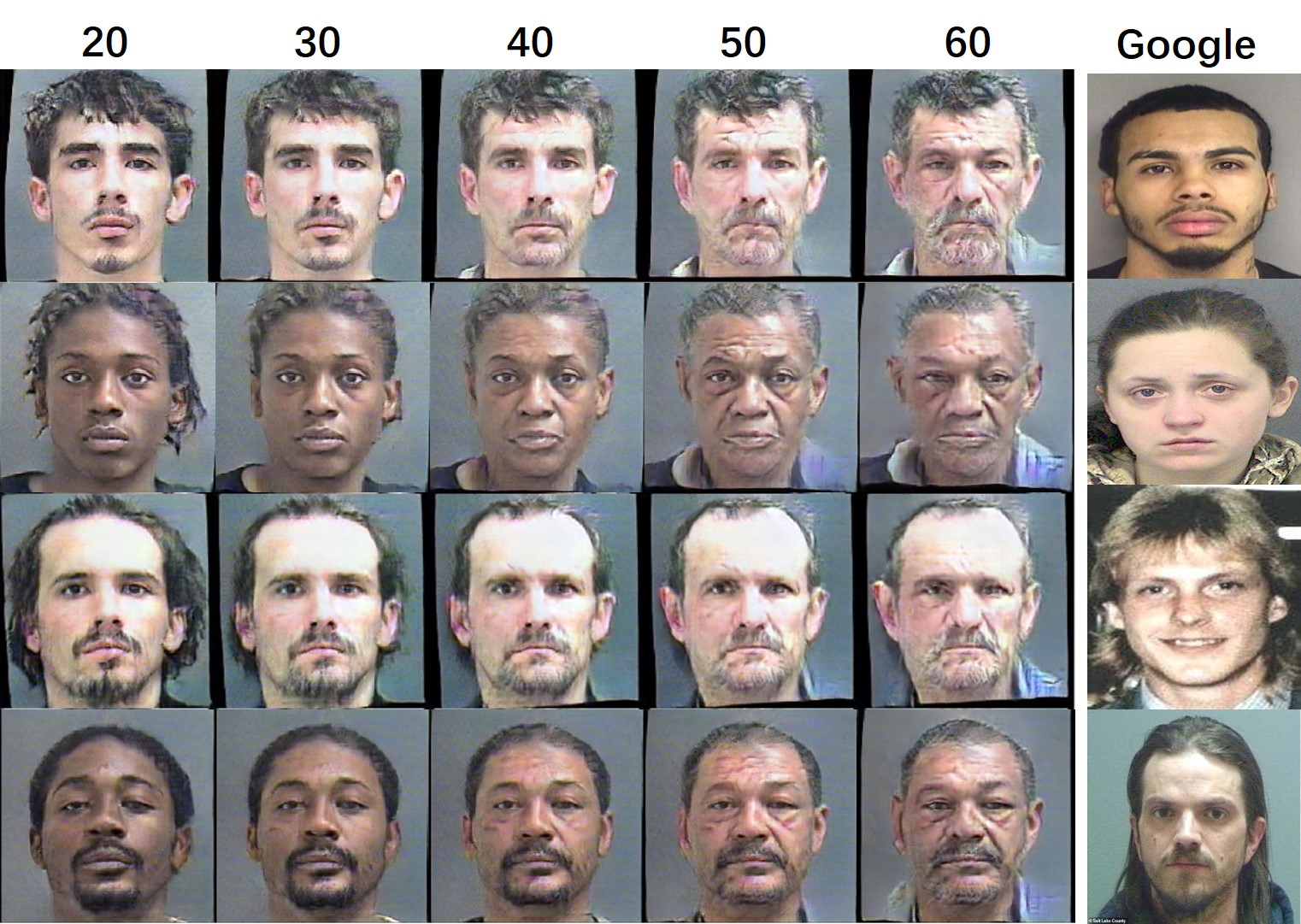}
\end{center}
\vspace{-0.3cm}
   \caption{Age generation results on Morph.  In the left, each row has the same age-irrelevant representation and each column has the same age-related representation. The right shows the most similar image from google for the generated image in the left.(All of the four subjects are generated from noise.)}\label{fig:morph_generation}
   \vspace{-0.3cm}
\end{figure}

On the other hand, by manipulating an arbitrary $\mu_R$ of age-related distribution and sampling both $\hat{z}_R$ and $\hat{z}_I$, UVA has the ability to generate faces with various ages. Fig. \ref{fig:cacd2000_generation} and \ref{fig:morph_generation} display the age generation results from 20 to 60 on the CACD2000 and Morph datasets, respectively. Each row has the same age-irrelevant representation $\hat{z}_I$ and each column has the same age-related representation $\hat{z}_R$. We can observe that the age-irrelevant information, such as gender and identity, is preserved across rows, and the age-related information, such as white hairs and wrinkles, is preserved across columns. These demonstrate that our UVA effectively disentangles age-related and age-irrelevant representations in the latent space. More results on the UTKFace and MegaAge-Asian datasets are presented in the supplementary materials, due to the page limitation.

\subsubsection{Generalized Abilities of UVA}

\begin{figure}[t]
\setlength{\abovecaptionskip}{0cm}
\begin{center}
\includegraphics[width=0.9\linewidth]{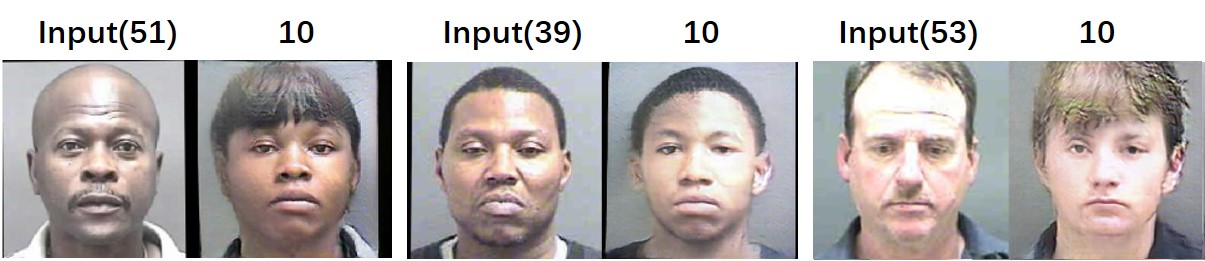}
\end{center}
\vspace{-0.3cm}
   \caption{Age translation results on Morph. Note that the target age (10 years old) is not existing in the Morph dataset.}\label{fig:nonexisting}
   \vspace{-0.3cm}
\end{figure}

Benefiting from the disentangling manner, UVA has the abilities to estimate the age-related distribution from the facial images, even if the training dataset is long-tailed. Since the facial images in Morph ages from 16 to 77, when performing age translation with 10 years old, as shown in Fig. \ref{fig:nonexisting}, it is amazing that UVA can synthesize photorealistic aging images. This observation demonstrates the generalized abilities of our framework, and also indicates that UVA can efficiently and accurately estimate the age-related distribution from the facial images, even if the training dataset performs a long-tailed age distribution.

\subsection{Quantitative Evaluation of UVA}
\subsubsection{Age Estimation}
To further demonstrate the disentangling ability of UVA, we conduct age estimation experiments on Morph and MegaAge-Asian, both of which are widely used datasets in age estimation task. We detail the evaluation metrics in the supplementary materials.

For experiment on Morph, following \cite{pan2018mean}, we choose the 80-20 random split protocol and report the mean absolute error(MAE).
The results are listed in Table \ref{tab:morph111}, where our model has not been pre-trained on the external data. We can see from Table \ref{tab:morph111} that the proposed UVA achieves comparative performance with the stat-of-the-art methods.

\begin{table}[htbp]\footnotesize
   \vspace{-0.2cm}
\centering
\caption{Comparisons with state-of-the-art methods on Morph. Lower MAE is better.}
\begin{threeparttable}
\begin{tabular}{lc|ccccclcc}
\hline
\multirow{2}{*}{Methods} & \multirow{2}{*}{Pre-trained} & Morph \\
\cline{3-3}
&&MAE\\
\hline
\hline
OR-CNN\cite{niu2016ordinal} & - & 3.34 \\
DEX\cite{rothe2018deep} & IMDB-WIKI$^ *$ & 2.68\\
Ranking \cite{chen2017using} &Audience& 2.96 \\
Posterior\cite{zhang2017quantifying} & - &2.87\\
SSR-Net\cite{yang2018ssr}&IMDB-WIKI &2.52\\
M-V Loss\cite{pan2018mean}&- &2.514\\
ThinAgeNet \cite{gaoage} &MS-Celeb-1M$^ *$ & 2.346\\
\hline
UVA&-&2.52\\
\hline
\end{tabular}\label{tab:morph111}
\begin{tablenotes}
        \footnotesize
        \item[*] \tiny{Used partial data of the dataset;}
      \end{tablenotes}
\end{threeparttable}
\vspace{-0.3cm}
\end{table}

For age estimation on MageAge-Asian, following \cite{zhang2017quantifying}, we reserve 3,945 images for testing and employ the Cumulative Accuracy (CA) as the evaluation metric. We report CA(3), CA(5), CA(7) as \cite{zhang2017quantifying,yang2018ssr} in Table \ref{tab:ca}. Note that our model has not pre-trained on the external data and achieves comparative performance.
\begin{table}[htbp]\footnotesize
   \vspace{-0.2cm}
\centering
\caption{Comparisons with state-of-the-art methods on MegaAge-Asian. The unit of CA(n) is $\%$. Higher CA(n) is better.}
\begin{threeparttable}
\begin{tabular}{lc|ccccc}
\hline
\multirow{2}{*}{Methods} & \multirow{2}{*}{Pre-trained} & \multicolumn{3}{c}{MegaAge-Asian}\\
\cline{3-5}
&&CA(3)&CA(5)&CA(7)\\
\hline
\hline
Posterior\cite{zhang2017quantifying} & IMDB-WIKI&  62.08& 80.43&90.42\\
MobileNet\cite{yang2018ssr}& IMDB-WIKI&44.0&60.6&-\\
DenseNet\cite{yang2018ssr}& IMDB-WIKI&51.7&69.4&-\\
SSR-Net\cite{yang2018ssr}& IMDB-WIKI&54.9&74.1&-\\
\hline
UVA& -&60.47    &79.95     &90.44\\
\hline
\end{tabular}\label{tab:ca}
\end{threeparttable}
\vspace{-0.3cm}
\end{table}

The age estimation results on Morph and MegaAge-Asian are nearly as good as the state-of-the-arts, which also demonstrates that the age-related representation is well disentangled from the age-irrelevant representation.
\subsubsection{Aging Accuracy}
 We conduct aging accuracy experiments of UVA on Morph and CACD2000, which is an essential quantitative metric to evaluate age translation. For fair comparison, following \cite{yang2017learning,li2018global}, we utilize the online face analysis tool of Face++ \cite{Face++} to evaluate the ages of the synthetic results, and divide the testing data of Morph and CACD2000 into four age groups: 30-(AG0), 31-40(AG1), 41-50(AG2), 51+(AG3). We choose AG0 as the input and synthesize images in AG1, AG2 and AG3. Then we estimate the ages of the synthesized images and calculate the average ages for each group.

\begin{table}[htbp]\footnotesize
\textbf{   \vspace{-0.2cm}}
\centering
\caption{Comparisons of the aging accuracy.}
\begin{tabular}{lccccccc}
\multicolumn{5}{c}{\small (a) on Morph} \\
\cline{1-5}
\cline{1-5}
\multirow{1}*{\textbf{Method}}&\multicolumn{1}{c}{\textbf{Input}}&\multicolumn{1}{c}{\textbf{AG1}}&\multicolumn{1}{c}{\textbf{AG2}}&\multicolumn{1}{c}{\textbf{AG3}}\\
\hline
\hline
CAAE\cite{zhang2017age} &-&28.13&32.50 &36.83\\
Yang et al.\cite{yang2017learning} &-&42.84 & 50.78  &    59.91\\
\cline{1-5}
UVA &-&36.59&50.14&60.78\\
\cline{1-5}
\textbf{Real Data} &28.19&38.89 & 48.10  &    58.22\\
\cline{1-5}
\multicolumn{5}{c}{\tiny} \\
\multicolumn{5}{c}{\small (b) on CACD2000} \\
\cline{1-5}
\cline{1-5}
\multirow{1}*{\textbf{Method}}&\multicolumn{1}{c}{\textbf{Input}}&\multicolumn{1}{c}{\textbf{AG1}}&\multicolumn{1}{c}{\textbf{AG2}}&\multicolumn{1}{c}{\textbf{AG3}}\\
\hline
\hline
CAAE\cite{zhang2017age} &-&31.32&    34.94 &     36.91\\
Yang et al.\cite{yang2017learning} &-&44.29 & 48.34  &    52.02\\
\cline{1-5}
UVA &-&39.19&45.73&51.32\\
\cline{1-5}
\textbf{Real Data} &30.73& 39.08&47.06   &53.68    \\
\cline{1-5}
\end{tabular}\label{tab:morph}
\vspace{-0.2cm}
\end{table}

As shown in Table \ref{tab:morph}, we compare the UVA with CAAE \cite{zhang2017age} and Yang et al. \cite{yang2017learning} on Morph and CACD2000. We observe that the transformed ages by UVA are closer to the natural data than by CAAE \cite{zhang2017age} and is comparable to Yang et al. \cite{yang2017learning}. Note that, for age translation among multiple ages, the proposed UVA only needs to train one model, but Yang et al. \cite{yang2017learning} need to train multiple models for each transformation, such as Input$\to$AG1, Input$\to$AG2 and Input$\to$AG3.

\subsubsection{Diversity}
In this subsection, we utilize the $Fr\acute{e}chet$ Inception Distance (FID) \cite{heusel2017gans} to evaluate the quality and diversity of the generated data. FID measures the $Fr\acute{e}chet$ distance between the generated and the real distributions in the feature space. Here, the testing images are generated from noise ${\hat{z}_I}$ with target ages ${\hat{z}_{R}}$, where ${\hat{z}_I}$ and ${\hat{z}_R}$ are sampled from $p\left( {{z_I}} \right)$ and $p\left( {{z_R}} \right)$, respectively. The FID results are detailed in Table \ref{tab:fid}.
\begin{table}[htbp]\footnotesize
   \vspace{-0.2cm}
\centering
\caption{The FID results on the four datasets. }
\begin{threeparttable}
\begin{tabular}{l|ccccccccc}
\hline
&Morph&CACD2000&UTKFace&MegaAge-Asian\\
\hline
FID & 17.154&  37.08 & 31.18 &32.79\\

\hline
\end{tabular}\label{tab:fid}
\end{threeparttable}
\vspace{-0.2cm}
\end{table}

\subsection{Ablation Study}
We conduct ablation study on Morph to evaluate the effectiveness of introspective adversarial learning and age preserving regularization. We introduce the details in the supplemental materials, due to the page limitation.

\section{Conclusion}

This paper has proposed a Universal Variational Aging framework for continuous age analysis. Benefiting from variational evidence lower bound, the facial images are disentangled into an age-related distribution and an age-irrelevant distribution in the latent space. A conditional introspective adversarial learning mechanism is introduced to improve the quality of facial aging synthesis. In this way, we can deal with long-tailed data and implement three different tasks, including age translation, age generation and age estimation. To the best of our knowledge, UVA is the first attempt to achieve facial age analysis in a universal framework. The qualitative and quantitative experiments demonstrate the superiority of the proposed UVA on five popular datasets, including CACD2000, Morph, UTKFace, MegaAge-Asian and FG-NET. This indicates that UVA can efficiently formulate the facial age prior, which contributes to both photorealistic and interpretable image synthesis for facial age analysis.
\clearpage
{\small
\bibliographystyle{ieee}
\bibliography{egbib}
}
\section{Supplementary Materials}
In this supplementary material, we first introduce the network architecture and the training algorithm of our proposed UVA. Then we depict the metrics used for age estimation experiments. Additional comparisons are conducted to demonstrate the continuous manner of UVA in Section 4. Besides, Section 5 presents the ablation study, followed by additional results of age generation and translation.

\subsection{Network Architecture}
The network architectures of the proposed UVA are shown in Table \ref{tab:UVA}. All images are aligned to 256$\times$256 as the inputs.
\begin{table}[htbp]\scriptsize
\centering
\caption{Network architectures of UVA.}
\begin{tabular}{l|l|l|l} \hline
Layer & Input  & Filter/Stride & Output Size\\
\hline
\multicolumn{4}{c}{Encoder} \\
\hline
E$\_$conv1 & $x$         & $5\times5 / 1$ & $256 \times 256 \times 16$\\
E$\_$avgpool1 & E$\_$conv1                & $2\times2 / 2$ & $128 \times 128 \times 16$\\
E$\_$res1 & E$\_$avgpool1                & $3\times3 / 1$ & $128 \times 128 \times 32$\\
E$\_$avgpool2  & E$\_$res1                & $2\times2 / 2$ & $64 \times 64 \times 32$\\
E$\_$res2  & E$\_$avgpool2                 & $3\times3 / 1$ & $64 \times 64 \times 64$\\
E$\_$avgpool3 & E$\_$res2                & $2\times2 / 2$ & $32 \times 32 \times 64$\\
E$\_$res3 & E$\_$avgpool3                & $3\times3 / 1$ & $32 \times 32 \times 128$\\
E$\_$avgpool4  & E$\_$res3                & $2\times2 / 2$ & $16 \times 16 \times 128$\\
E$\_$res4  & E$\_$avgpool4                 & $3\times3 / 1$ & $16 \times 16 \times 256$\\
E$\_$avgpool5 & E$\_$res4                & $2\times2 / 2$ & $8 \times 8 \times 256$\\
E$\_$res5 & E$\_$avgpool5                & $3\times3 / 1$ & $8 \times 8 \times 512$\\
E$\_$avgpool6 & E$\_$res5                & $2\times2 / 2$ & $4 \times 4 \times 512$\\
E$\_$res6 & E$\_$avgpool6                & $3\times3 / 1$ & $4 \times 4 \times 512$\\
E$\_$flatten, fc & E$\_$res6               & - & 1024\\
E$\_$split  & E$\_$fc &-&  \scriptsize 256,256,256,256\\
E$\_$repara1 & E$\_$split[0,1] &-& 256\\
E$\_$repara2 & E$\_$split[2,3] &-& 256\\
E$\_$cat & E$\_$repara1, E$\_$repara2,                & - & 512\\

\hline
\multicolumn{4}{c}{Generator} \\
\hline
G$\_$fc,reshape & E$\_$cat                & - & $4 \times 4 \times 512$\\
G$\_$res1 & G$\_$fc         & $3\times3 / 1$ & $4 \times 4 \times 512$\\
G$\_$upsample1 & G$\_$res1                & $\times2$, nearest & $8 \times 8 \times 512$\\
G$\_$res2 & G$\_$upsample1                & $3\times3 / 1$ & $8 \times 8 \times 512$\\
G$\_$upsample2 & G$\_$res2                 & $\times2$, nearest & $16 \times 16 \times 512$\\
G$\_$res3       & G$\_$upsample2            & $3\times3 / 1$ & $16 \times 16 \times 256$\\
G$\_$upsample3 & L$\_$res3                & $\times2$, nearest & $32 \times 32 \times 256$\\
G$\_$res4 & G$\_$upsample3                & $3\times3 / 1$ & $32 \times 32 \times 128$\\
G$\_$upsample4 & G$\_$res4                 & $\times2$, nearest & $64 \times 64 \times 128$\\
G$\_$res5       & G$\_$upsample4            & $3\times3 / 1$ & $64 \times 64 \times 64$\\
G$\_$upsample5 & L$\_$res5                & $\times2$, nearest & $128 \times 128 \times 64$\\
G$\_$res6 & G$\_$upsample5                & $3\times3 / 1$ & $128 \times 128 \times 32$\\
G$\_$upsample6 & G$\_$res7                 & $\times2$, nearest & $256 \times 256 \times 32$\\
G$\_$res7       & G$\_$upsample6            & $3\times3 / 1$ & $256 \times 256 \times 32$\\
G$\_$conv1       & G$\_$res7            & $5\times5 / 1$ & $256 \times 256 \times 3$\\

\hline

\end{tabular}\label{tab:UVA}
\end{table}
\subsection{Training Algorithm}
The training process of UVA is described in Algorithm \ref{alg:training}.

\begin{algorithm}[t]\scriptsize
\setstretch{1.3}
\caption{Training UVA Model}
\label{alg:training}
\begin{algorithmic}[1]
\State \(\theta_G, \phi_E \gets\) Initialize~ network~ parameters
\While{not converged}
\State \(X \gets\) Random mini-batch from training set
\State \(\mu_R, \sigma _R, \mu_I, \sigma_I \gets\) \(Enc(X)\)
\State \(Z_R \gets\) \(\mu _R + \epsilon\odot \sigma _R\), \(Z_I \gets\) \(\mu _I + \epsilon\odot \sigma _I\)
\State \({{\mathord{\buildrel{\lower3pt\hbox{$\scriptscriptstyle\frown$}} \over Z} }_I}\) Samples from prior \(N(0,I)\)
\State \(X_r \gets\) \(Dec(Z_R, Z_I)\), \(X_s \gets\) \(Dec(Z_R, {{\mathord{\buildrel{\lower3pt\hbox{$\scriptscriptstyle\frown$}} \over Z} }_I})\)
\State \(Z_R' \gets\) \(\mu _R' + \epsilon\odot \sigma _R'\), \(Z_I' \gets\) \(\mu _I' + \epsilon\odot \sigma _I'\)

\State \(Z_R'' \gets\) \(\mu _R'' + \epsilon\odot \sigma _R''\), \(Z_I'' \gets\) \(\mu _I'' + \epsilon\odot \sigma _I''\)
\State \( L_{rec} \gets L_{rec}(X_r, X) \)
\State \( L_{age\_{kl}} \gets L_{age_{kl}}(\mu_R, \sigma _R, Y) \)
\State \( L_{reg} \gets |\mu_R- Y| \)
\State \(\mu_R', \sigma _R', \mu_I', \sigma_I' \gets\) \(Enc(X_r), \mu_R'', \sigma _R'', \mu_I'', \sigma_I'' \gets\) \(Enc(X_s)\)
\State \( L_{age\_keep} \gets |\mu_R'- Y|+ |\mu_R''- Y|\)
\State \( L^{E}_{adv} \gets L_{kl}(\mu_I, \sigma_I) +  \alpha\{[m - L_{kl}(\mu_I', \sigma_I')]^+ + [m - L_{kl}(\mu_I'', \sigma_I'')]^+\}    \)
\State \(L^{G}_{adv} \gets  \alpha\{L_{kl}(\mu_I', \sigma_I') + L_{kl}(\mu_I'', \sigma_I'')\}  \)
\State \( \phi_E \gets \phi_E - \eta\nabla_{\phi_E} (  L_{rec} + \lambda _1 L_{age\_kl} + \lambda _2 L^{E}_{adv} +  \lambda _3 L_{reg}) \)
 \;\;\;\;\;\;\;\;\;\;\;\;\;\;\;\;\;\;\;\;\;\;\ \Comment{Perform Adam updates for \(\phi_E\)}
\State \(\theta_G \gets \theta_G - \eta\nabla_{\theta_G} (L_{rec} + \lambda _4 L^{G}_{adv} + \lambda _5 L_{age\_keep} ) \)
 \;\;\;\;\;\;\;\;\;\;\;\;\;\;\;\;\;\;\;\;\;\;\ \Comment{Perform Adam updates for \(\theta_G\)}
\EndWhile
\end{algorithmic}
\end{algorithm}

\subsection{Evaluation Metrics of Age Estimation}
We evaluate the performance of UVA on age estimation task with the Mean Absolute Error(MAE) and Cumulative Accuracy (CA).

MAE is widely used to evaluate the performance of age estimation models,which is defined as the average distance between the inferred age and the ground-truth:
\begin{equation}
\begin{array}{c}
MAE = \frac{1}{N}\sum\limits_{i = 1}^N {|{{\hat y}_i} - {y_i}|}
\end{array}
\end{equation}

The Cumulative Accuracy (CA) is defined as:
\begin{equation}
\begin{array}{c}
CA\left( n \right) = \frac{{{K_n}}}{K} \times 100\%
\end{array}
\end{equation}
where ${K_n}$ is the number of the test images whose absolute estimated error is smaller than $n$.

\subsection{Additional Comparison Results on FG-NET}
\begin{figure}[h]
\setlength{\abovecaptionskip}{0cm}
\begin{center}
\includegraphics[width=1\linewidth]{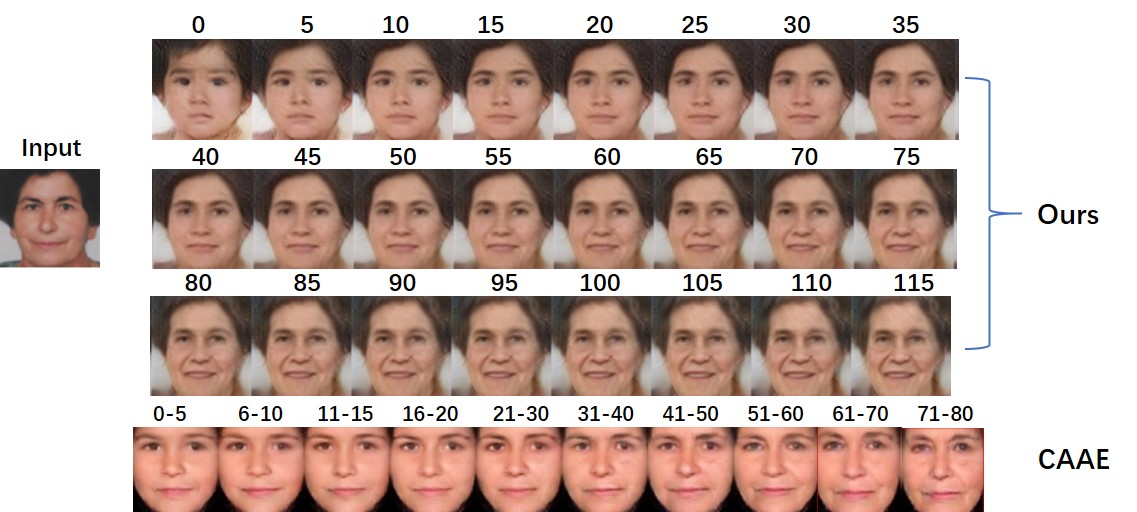}
\end{center}
\vspace{-0.3cm}
   \caption{Comparison with CAAE on FG-NET.}\label{fig:compare_caae}
   \vspace{-0.3cm}
\end{figure}
Additional comparison results with the conditional adversarial autoencoder (CAAE) \cite{zhang2017age} on FG-NET are shown in Figure \ref{fig:compare_caae}. Note that our model is trained on UTKFace. We observe that CAAE can only translate the input into specific age groups, including 0-5, 6-10, 11-15, 16-20, 21-30, 31-40, 41-50, 51-60, 61-70, and 71-80, while UVA can perform age translation with the arbitrary age. In addition, CAAE only focuses on the face appearances without hair, while UVA achieves age translation on the whole face. As shown in Fig. \ref{fig:compare_caae}, the hairline gradually becomes higher as age increases.

\subsection{Ablation Study}
\begin{figure}[h]
\setlength{\abovecaptionskip}{0cm}
\begin{center}
\includegraphics[width=1\linewidth]{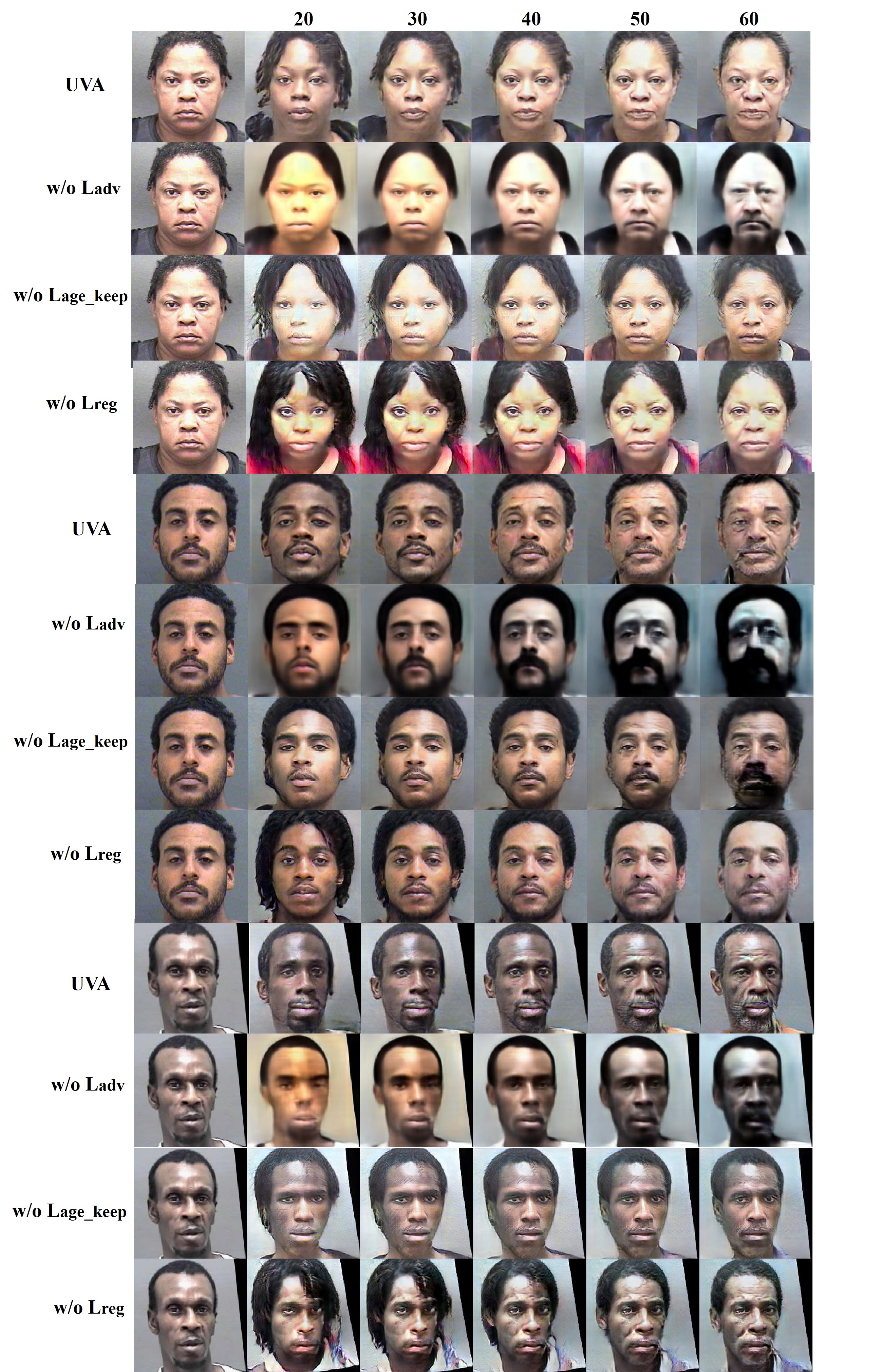}
\end{center}
\vspace{-0.3cm}
   \caption{Model comparison: age translation results of UVA and its variants. For each subject, the first row is the synthesis results of UVA, while the second, third and fourth rows are the synthesis results without introspective adversarial loss, age preserving loss and age regularization term, respectively.}\label{fig:ablation}
   \vspace{-0.0cm}
\end{figure}
In this section, we conduct ablation study on Morph to evaluate the effectiveness of the introspective adversarial loss, the age preserving loss and the age regularization term, respectively. We report the qualitative visualization results for a comprehensive comparison.
Fig. \ref{fig:ablation} shows visual comparisons between UVA and its three incomplete variants. We observe that the proposed UVA is visually better than its variants across all ages. Without the $L_{age\_keep}$ or $L_{reg}$ loss, the aging images lack the detailed texture information. Besides, without the $L_{adv}$ loss, the aging faces are obviously blur.  The visualization results demonstrate that all three components are essential for UVA.

\subsection{Age Generation}

\begin{figure}[h]
\setlength{\abovecaptionskip}{0cm}
\begin{center}
\includegraphics[width=1\linewidth]{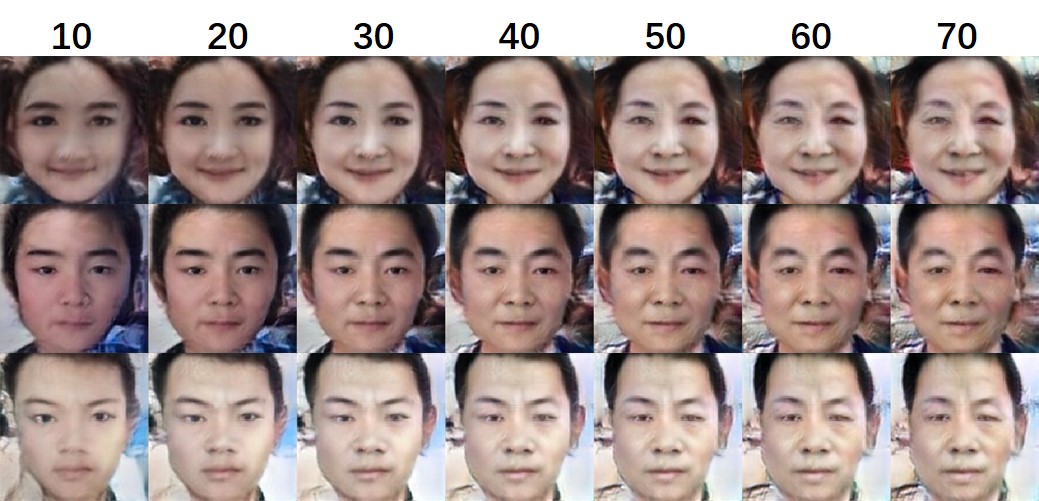}
\end{center}
\vspace{-0cm}
   \caption{Age generation results on MegaAge-Asian. Each row has the same age-irrelevant representation and each column has the same age-related representation. The images in MegaAge-Asian are low-resolution and limited-quality, leading to blurred synthesized results.}\label{fig:generation_ma}
   \vspace{-0cm}
\end{figure}

\begin{figure*}[h]
\setlength{\abovecaptionskip}{0cm}
\begin{center}
\includegraphics[width=0.9\linewidth]{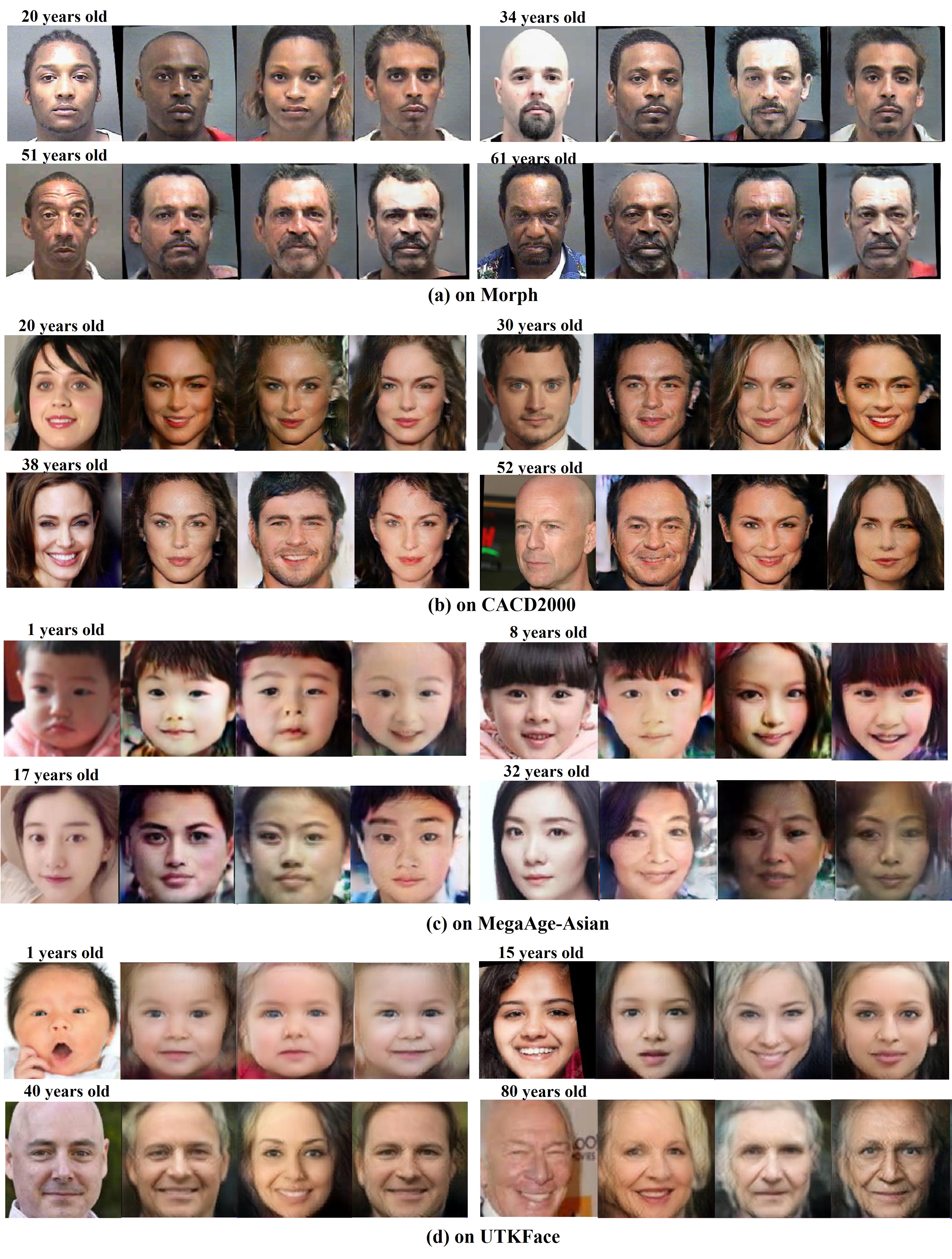}
\end{center}
\vspace{-0.3cm}
   \caption{Age generation results with age feature extracted from the given image on Morph, CACD2000, MegaAge-Asian and UTKFace. For each subject, the first column is the input face, while the rest three columns are the generated faces.}\label{fig:generation}
   \vspace{-0.3cm}
\end{figure*}

Benefiting from the disentangled age-related and age-irrelevant distributions in the latent space, the proposed UVA is able to realize age generation.
On the one hand, given a facial image, when fixing age-related distribution $z_R$ and sampling $\hat{z}_I$ from noise, UVA can generate various faces with the specific age. Fig. \ref{fig:generation} shows the age generation results on Morph, CACD2000, MegaAge-Asian and UTKFace.
On the other hand, by manipulating $\mu_R$ of the age-related distribution and sampling $\hat{z}_I$ from noise, UVA can generate faces with various ages. Fig. \ref{fig:generation_ma} shows the age generation results from 10 to 70 on MegaAge-Asian.

\subsection{Age Translation}
\begin{figure}[h]
\setlength{\abovecaptionskip}{0cm}
\begin{center}
\includegraphics[width=1\linewidth]{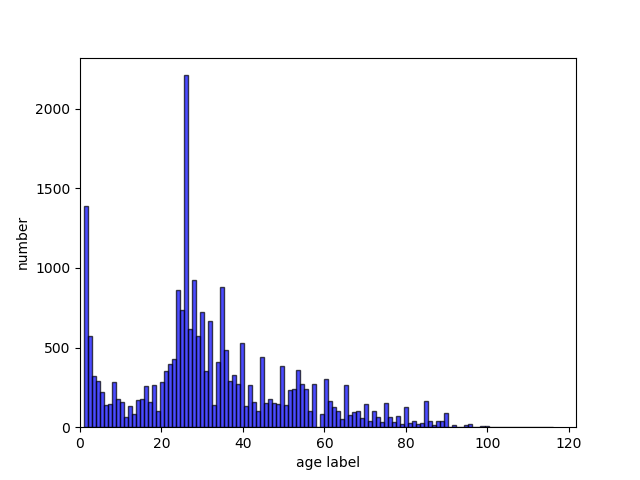}
\end{center}
\vspace{-0.3cm}
   \caption{Age distribution in UTKFace.}\label{fig:long-tailed}
   \vspace{-0.3cm}
\end{figure}

\begin{figure*}[t]
\setlength{\abovecaptionskip}{0cm}
\begin{center}
\includegraphics[width=1\linewidth]{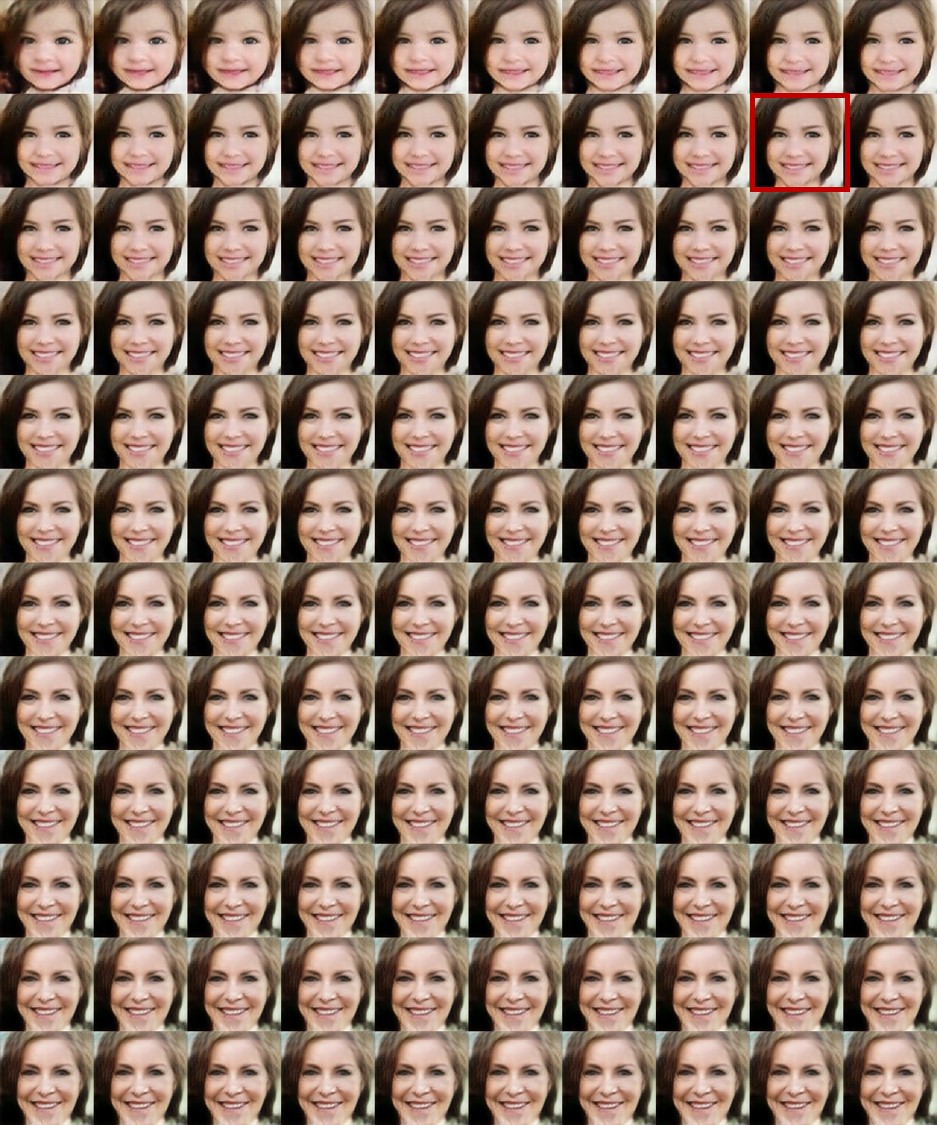}
\end{center}
\vspace{-0.2cm}
   \caption{Age translation results on UTKFace from 0 to 119 years old with 1-year-old interval. The image in the red box is the input (18 years old). The other images are the age translation results from 0 to 119 years old. }\label{fig:generation_utkface2}
   \vspace{-0.2cm}
\end{figure*}

\begin{figure*}[t]
\setlength{\abovecaptionskip}{0cm}
\begin{center}
\includegraphics[width=1\linewidth]{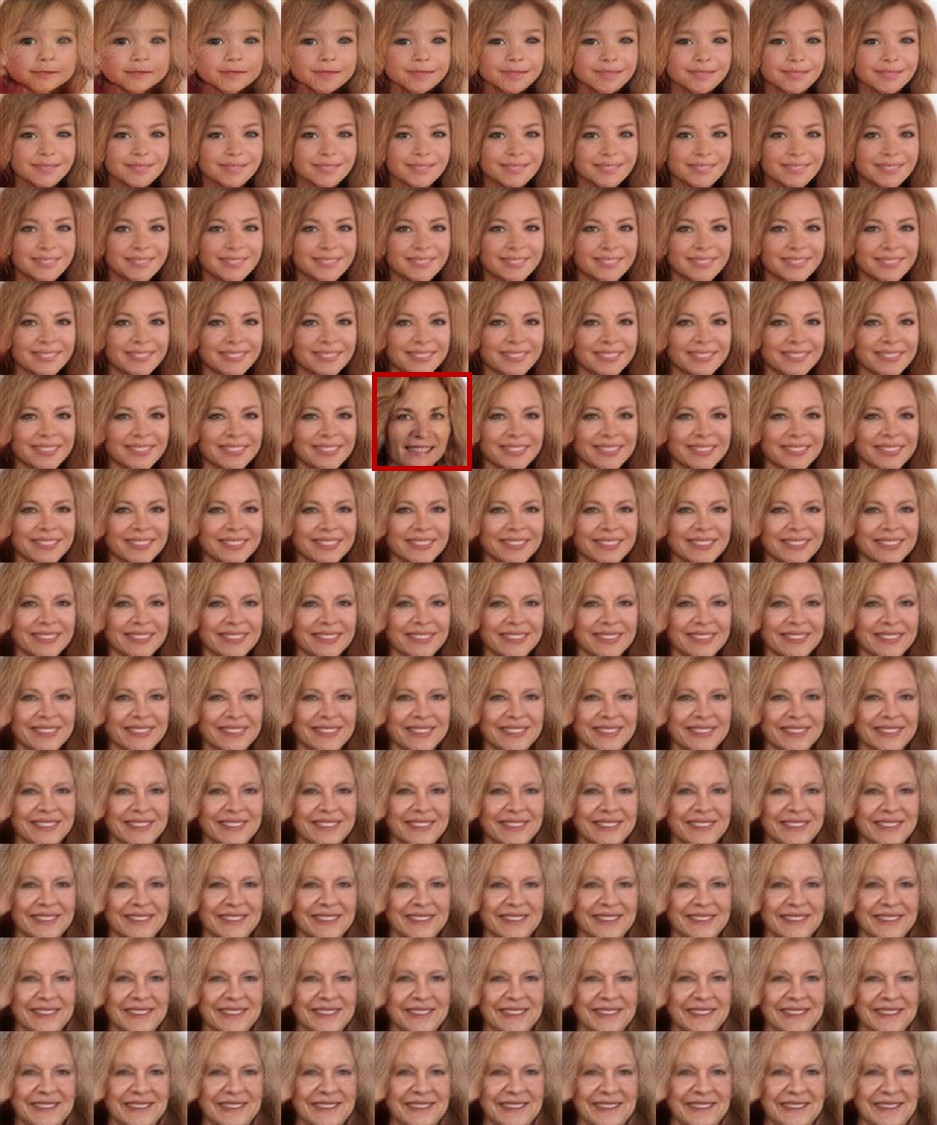}
\end{center}
\vspace{-0.2cm}
   \caption{Age translation results on UTKFace from 0 to 119 years old with 1-year-old interval. The image in the red box is the input (44 years old). The other images are the age translation results from 0 to 119 years old. }\label{fig:generation_utkface3}
   \vspace{-0.2cm}
\end{figure*}

\begin{figure*}[t]
\setlength{\abovecaptionskip}{0cm}
\begin{center}
\includegraphics[width=1\linewidth]{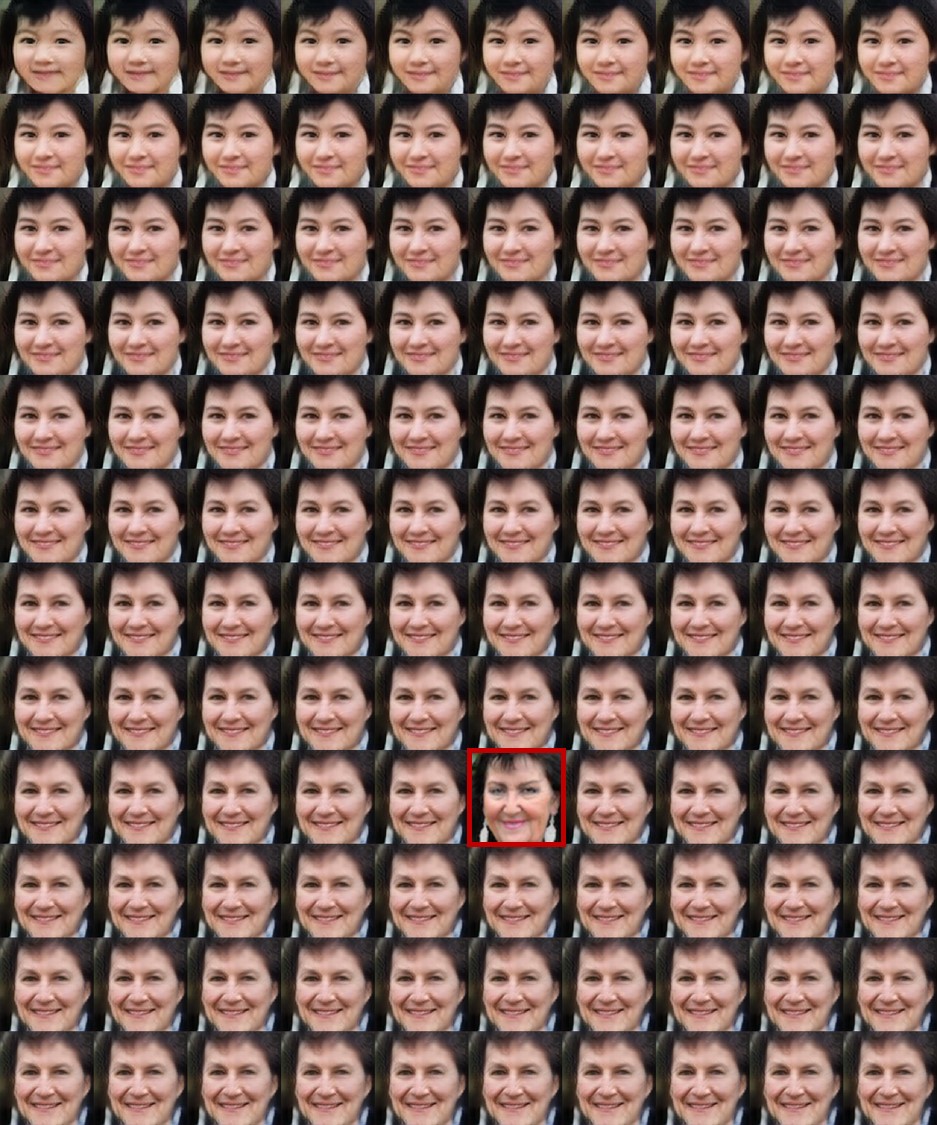}
\end{center}
\vspace{-0.2cm}
   \caption{Age translation results on UTKFace from 0 to 119 years old with 1-year-old interval. The image in the red box is the input (85 years old). The other images are the age translation results from 0 to 119 years old. }\label{fig:generation_utkface4}
   \vspace{-0.2cm}
\end{figure*}

As shown in Fig. \ref{fig:long-tailed}, we observe that UTKFace exhibits a long-tailed age distribution.
The age translation results on UTKFace are presented in Fig. \ref{fig:generation_utkface2}, \ref{fig:generation_utkface3} and \ref{fig:generation_utkface4}. The ages range from 0 to 119 years old with 1-year-old aging interval. Since the images in UTKFace range from 0 to 116 years old, it is amazing that UVA can synthesize photo-realistic aging images even with 117,118 and 119 years old.
These results suggest that UVA can effeciently and accurately estimate the age-related distribution from the facial images, even if the training dataset performs a long-tailed age distribution. Considering that generating high-resolution images is significant to enlarge the application field of face aging, in the future, we will build a new facial age dataset to support the age-related study.


\end{document}